\documentclass[10pt,journal,doublecolumn]{IEEEtran}
\usepackage{amsmath,amssymb,threeparttable,placeins,enumerate,caption}
\usepackage{mathtools,relsize,cite}
\usepackage[pdftex]{graphicx}
\usepackage{epstopdf}
\usepackage[usenames]{color}
\usepackage{amsfonts}
\usepackage{latexsym}
\usepackage{bm}
\usepackage{subfigure,multirow,makecell,stfloats}

\definecolor{burntorange}{rgb}{0.8, 0.33, 0.0}

\usepackage{algorithm}
\usepackage[noend]{algpseudocode}

\newtheorem{Theorem}{{{\textit{Theorem}}}}

\newtheorem{Definition}{{{\textit{Definition}}}}

\newtheorem{Remark}{{{\textit{Remark}}}}

\newtheorem{Example}{{{\textit{Example}}}}

\begin{document}
	
	
	\title{HpGAN: Sequence Search with Generative Adversarial Networks}
	\author{Mingxing Zhang, Zhengchun Zhou, Lanping Li, Zilong Liu,  Meng Yang, and Yanghe Feng
\thanks{M. Zhang is with the School of Information Science and Technology, Southwest Jiaotong University,
Chengdu, 610031, China  (e-mail: mingxingzhang@my.swjtu.edu.cn)}
\thanks{Z. Zhou and M. Yang are with the School of Mathematics, Southwest Jiaotong University,
Chengdu, 610031, China  (e-mail:
zzc@swjtu.edu.cn, mekryang@gmail.com)}
\thanks{L. Li is with the School of Electrical Engineering and Information, Southwest Petroleum University,
Chengdu, 610500, China  (e-mail: lilanping523@gmail.com)}
\thanks{Z. Liu is with School of Computer Science and Electronic Engineering, University of Essex, UK.(email: zilong.liu@essex.ac.uk)}
\thanks{Y. Feng is with the College of Systems Engineering, National University of Defense Technology,
Changsha, 410073, China. (e-mail: fengyanghe@nudt.edu.cn)}}
\maketitle
	
	\begin{abstract}
Sequences play an important role in many engineering applications and systems. Searching sequences with desired properties has long been an interesting but also challenging research topic. This article proposes a novel method, called HpGAN, to search desired sequences algorithmically using generative adversarial networks (GAN). HpGAN is based on the idea of zero-sum game to train a generative model, which can generate sequences with characteristics similar to the training sequences. In HpGAN, we design the Hopfield network as an encoder to avoid the limitations of GAN in generating discrete data. Compared with traditional sequence construction by algebraic tools, HpGAN is particularly suitable for intractable problems with complex objectives which prevent mathematical analysis. We demonstrate the search capabilities of HpGAN in two applications: 1) HpGAN successfully found many different mutually orthogonal complementary code sets (MOCCS) and optimal odd-length Z-complementary pairs (OB-ZCPs) which are not part of the training set. In the literature, both MOCSSs and OB-ZCPs have found wide applications in wireless communications. 2) HpGAN found new sequences which achieve four-times increase of signal-to-interference ratio---benchmarked against the well-known Legendre sequence---of a mismatched filter (MMF) estimator in pulse compression radar systems. These sequences outperform those found by AlphaSeq (IEEE Trans. Neural Netw. Learn. Syst. 31(9) (2020) 3319--3333).
	\end{abstract}

\vspace{-0.2cm}
	
\section{Introduction}
	\IEEEPARstart{A}{sequence} is a list of elements arranged in a specific order. Good sequences form a core component of many information systems. As wireless mobile communication technologies evolving rapidly meet the increasingly stringent and diverse requirements of various data services, it is critical to design sequences with different properties. For example, sequences with low autocorrelation are widely used in pulse compression radars, sonars and channel synchronization of digital communication \cite{Mertens1996}. Also, orthogonal sequence sets are used to distinguish signals from different users in  cellular code-division multiple access (CDMA) systems \cite{Viterbi1995}.

In the open literature, sequences are commonly designed by algebraists using mathematical tools such as finite field theory, algebraic number theory, and
character theory. However, in practical scenarios, it may be difficult to construct sequences using specific mathematical tools. To facilitate the use of mathematical tools, some constraints in terms of the sequence lengths, alphabet size, and set size may be imposed. For example, Davis and Jedwab constructed polyphase complementary sets of sequences by generalized Boolean function, but the sequence lengths are restricted to $2^m$, where $m$ is nonnegative integer numbers \cite{James1999}.

Algorithm design is an another direction of obtaining good sequences which may not be found by algebraic tools. A key issue here is whether good sequences can be found within a reasonable time by algorithms.
There are two major types of sequence search algorithms: one is to design optimal algorithms through specific mathematical analysis, and the other is by heuristic algorithms inspired by, for example, natural laws. A good optimization algorithm can effectively find sequences with guaranteed convergence \cite{Stoica2009, Soltanalian2012, Song2015, Kerahroodi2017}. However, such optimization algorithms need to be carefully designed case by case and their derivations may be not straightforward. Heuristic algorithms are not so sensitive to specific problems and can handle complex optimization problems efficiently, such as simulated annealing (SA) \cite{Deng1993, Deng2004}, evolutionary algorithm (EA) \cite{Deng1999, Mow2015} and neural network \cite{Hu1997}.

In recent years, as artificial intelligence has achieved great success in various fields \cite{Hinton2006, LeCun2015}, a tremendous research attention has been paid to different neural networks.  In the communication, researchers have used deep learning and reinforcement learning (RL) to allocate spectrum to avoid mutual interference \cite{Li2010, Faganello2013, Wang2018, Chang2018, Naparstek2018, Liu2020}. In 2020, based on the basic framework of AlphaGo, Shao \emph{et al.} proposed a new reinforcement learning model AlphaSeq to discover sequences \cite{Shao2020}. AlphaSeq treats the sequence discovery problem as an episodic symbol-filling game, in which a player fills symbols in the vacant positions of a sequence set sequentially during an episode of the game. However, in reinforcement learning, it is necessary to provide experience to the agent through extensive exploration, making AlphaSeq time-consuming during the sequence search. In particular, it may be difficult for AlphaSeq to search for longer length sequences.

Recently, generative adversarial networks (GAN), which is a class of generative model, have been proposed by Goodfellow \emph{et al.} \cite{Goodfellow2014}. Inspired by the two-person zero-sum game in game theory, GAN consists of a generative net $G$ and  discriminative net $D$. Specifically, the generative net $G$ is used to capture the data distribution, whilst the discriminative net $D$ helps estimate the probability that a sample came from the training data rather than $G$.  This approach has been widely applied in computer vision for generating samples of natural images \cite{Denton2015}.

That being said, GAN is designed to generate real-valued continuous data, and may have some limitations when dealing with discrete data. The reason is that the discrete output of the generative model makes it difficult to transfer the gradient update from the discriminant model to the generative model \cite{Huszar2015, Yu2017}. In \cite{Yu2017}, Yu \emph{et al.} proposed a generative model SeqGAN that can generate discrete data. The SeqGAN directly performs gradient policy updates to bypass the generator differentiation problem. This is done by combining RL and GAN, using the output of $D$ as a reward for RL, and then updating $G$ with the policy gradient of RL. SeqGAN has achieved remarkable successes in natural language (such as speech
language and music generation) by using recurrent neural networks (RNN) as the generator $G$. Similar to AlphaSeq, SeqGAN generates a sequence by symbol filling, where a symbol is generated each time with RNN guessing. However, the discriminative net $D$ can only assess a complete sequence, while for a partially generated sequence, it is essential to balance its current score and the future one once the entire sequence has been generated. Thus, to evaluate the score for an intermediate symbol, the authors applied Monte Carlo search to sample the unknown last remaining symbols. As a result, a high computational complexity may be inevitable.

In this paper, to find the sequences which may be difficult to generate with systematic constructions, we propose a new network architecture called HpGAN. In HpGAN, we design the encoder and decoder through the Hopfield network \cite{Hopfield1982}, \cite{Hopfield1985}, where the encoder can convert discrete data into continuous data, and the decoder can restore continuous data to discrete data. The encoding module has two main functions: one is to enable GAN produce continuous data; the other is to solve the problem of small samples faced by GAN. Both the  generative net $G$ and the  discriminative net $D$ in HpGAN are multilayer perceptrons, which are easy to implement with lower computational complexity than other networks, such as convolutional neural networks (CNN).

We demonstrate the capability of HpGAN for the search of the following two types of seqences:
\begin{enumerate}
  \item
  We use HpGAN to find some MOCCSs, which are different from the training data. Complementary pairs were proposed by Golay in his work on infrared spectrometry in 1949 \cite{Golay1949}, where their mathematical properties were studied in 1961 \cite{Golay1961}. the existing known binary GCPs only have even-lengths in the form of $2^{\alpha}10^{\beta}26^{\gamma}$ where $\alpha, \beta, \gamma$ are nonnegative integers \cite{Fan1996,Parker2002}. Motivated by the limited admissible lengths of binary GCPs, Fan \emph{et al.} proposed ``Z-complementary pair (ZCP)'' which features zero aperiodic auto-correlation sums for certain out of-phase time-shifts around the in-phase position \cite{Fan2007}. Such a region is called a zero-correlation zone (ZCZ). To further demonstrate the effectiveness of HpGAN, we also search odd-length binary ZCP (OB-ZCP) \cite{Liu2014}, where the optimal ones display the closest correlation property to that of GCPs. Specifically, an optimal OB-ZCP has the maximum possible ZCZ width of $(N+1)/2$ and minimum possible out of-zone aperiodic autocorrelation sum magnitude of $2$, where $N$ denotes the sequence length (odd).

  Up to date, GCPs and OB-ZCPs have found numerous applications in wireless engineering such as radar sensing \cite{Budisin1985}, channel estimation \cite{Wang2007}, synchronization in 3G standard \cite{Popovic2003}, and peak-to-mean envelope power ratio (PMEPR) reduction \cite{Popovic1991, Davis1999}. The concept of complementary pairs was extended later to complementary sets of sequences (CSS)\cite{Tseng1972}, which are widely used in  multi-carrier code-division multiple access (MC-CDMA) systems to eliminate multi-path interference and multi-user interference \cite{Liu2015}, \cite{Sun2015}.

  In this application, we demonstrate that HpGAN can reach the global optimal value by in finding MOCCSs and OB-ZCPs. We use the sequences obtained by systematic constructions as the training data, and then use HpGAN to generate many optimal sequences that are different from the training data.

  \item We use HpGAN to find new phase-coded sequences superior to the known sequences for pulse compression radar systems. In modern complex applications, the radar is required to have a large detection range and high resolution. However, for a pulsed radar system that transmits a fixed carrier frequency, its resolution is inversely proportional to the transmitted pulse width. Pulse compression radar can take into account the detection range and resolution at the same time by modulated pulses \cite{Shao2020, Boehmer1967, Davis2007}.
      The performances of the phase-coded sequences obtained by HpGAN outperform that from the existing known sequences, such as Legendre sequence \cite{Golay1983}, and AlphaSeq sequence \cite{Shao2020}. Specifically, compared with Legendre sequence, sequences found by HpGAN improve the signal-to-noise ratio (SIR) of the mismatched filter (MMF) estimator by four times. Since optimal sequence of MMF is unknown, especially for large-length sequences, we exploit this application to verify the search ability of HpGAN. The sequences found by HpGAN is shown to be superior to the training data, which are obtained by stochastic search algorithms. This shows that the performance of HpGAN is not limited by the training set.
\end{enumerate}

The remainder of this paper is organized as follows. Section II formulates the sequence search problem and outlines the framework of HpGAN. Sections III and IV present the applications of HpGAN for  MOCCSs and phase-coded sequences, respectively. Section V concludes this paper. Throughout this article, lowercase bold letters denote vectors and uppercase bold letters denote matrices.

\section{Methodology}
\subsection{Problem Formulation}
Let $\mathcal{C}=\{\bm{x}_{1},\bm{x}_{2},\ldots,\bm{x}_{M}\}$ denote a binary sequence set which consists of $M$ different sequences of the same length $N$, where the $k$th sequence is given by $\bm{x}_{k}=[x_{1},x_{2},\ldots,x_{N}]^{T}$ with $x_{i}=\{-1,1\}$ for all $1\leq i \leq N$. Let $\mathcal{M(C)}$ be a measure of the goodness of sequence set $\mathcal{C}$. Our objective is to find a sequence set $\mathcal{C}^{*}$ with the best $\mathcal{M}^{*}$ with certain criteria. For example, an optimal sequence set whose maximum crosscorrelation meets the celebrated Welch bound is desired for mitigation of multiuser interference when it is applied in code-division multiple access. Exhaustive search of optimal binary sequence sets may be infeasible due to the prohibitively high complexity of  $O(2^{MN})$.

Let $P_{\mathcal{C}}$ be the probability distribution function of all sequence sets with $\mathcal{M(C)}\leq \xi$, i.e., $P_{\mathcal{C}}=P(\mathcal{M(C)}\leq \xi)$, where $\mathcal{C}\in\Omega$ and $\Omega$ denotes be the solution space of all binary sequence sets $\mathcal{C}$. The problem of sequence search can be transformed into the task of training a generative model $G$ to learn the probability distribution function $P_{\mathcal{C}}$. By doing so, new sequence sets can be obtained with probability distribution function $P_{G}$. We adopt the idea of GAN by regarding the task of learning the probability distribution function $P_{\mathcal{C}}$ as a zero-sum game. The core idea of GAN is based on Nash equilibrium in game theory. It sets the two parties involved in the game as a generator and a discriminator. The purpose of the generator is to learn the real data distribution $P_{\mathcal{C}}$ as much as possible, while the purpose of the discriminator is to judge whether the input data comes from $P_{\mathcal{\mathcal{C}}}$ or $P_{G}$ as much as possible. In order to win the game, these two players need to constantly optimize and improve their generating ability and discriminating ability, respectively, until a  Nash equilibrium is attained.
\subsection{Methodology}
It is necessary to obtain a large amount of training data when searching sequences by GAN. It is easy to generate sequences with good rather than the best $\mathcal{M(C)}$ as training data by existing algorithms or constructions, such as stochastic search methods. Unfortunately, the application of GAN suffers from two problems. Firstly, GAN is designed for generating real-valued, continuous data but has difficulties in directly generating sequences of discrete data, such as discrete phase sequences. This is because in GANs, the generator starts with random sampling, followed by a determistic transform, according to the model parameters. As such, the gradient of the loss from discriminant model $D$ with respect to the outputs by $G$ is used to guide the generative model $G$ (paramters) to slightly change the generated value to make it more realistic. If the generated data is based on discrete phases, the ``slight change'' guidance from the discriminative net makes little sense because there may be no corresponding phase for such slight change in the limited dictionary space. Secondly, it is difficult for GAN to learn the real distribution, when there is a small amount of training data. Roughly speaking, the more sampled data we have, the closer the learned probability distribution approaching to the true one. However, even with the existing algorithms or constructions, the available training datasets are still very limited.

In this paper, to address the above two difficulties, we develop a new network framework called HpGAN by combining Hopfield network and GAN, in which we design an encode method to map binary sequences to a continuous interval. Then we exploit discrete Hopfield network framework to decode the generated continuous datasets into the binary ones when GAN reaches equilibrium. The overall algorithmic framework of HpGAN as shown in Fig. 1. First of all, we generate a dataset through the existing algorithm or structure as the initial sample dataset, compile the discrete dataset into continuous dataset by using the encoder, and then feed the encoded dataset into GAN. A generator $G$ which can generate data similar to the training data is obtained when the GAN reaches the Nash equilibrium. Secondly, the generator obtains a large amount of data, and selects some good datasets as new samples to update the initial samples. Through the iterative process, GAN can progressively generate better dataset than the initial samples.
\begin{figure}
  \centering
  \includegraphics[width=3.0in]{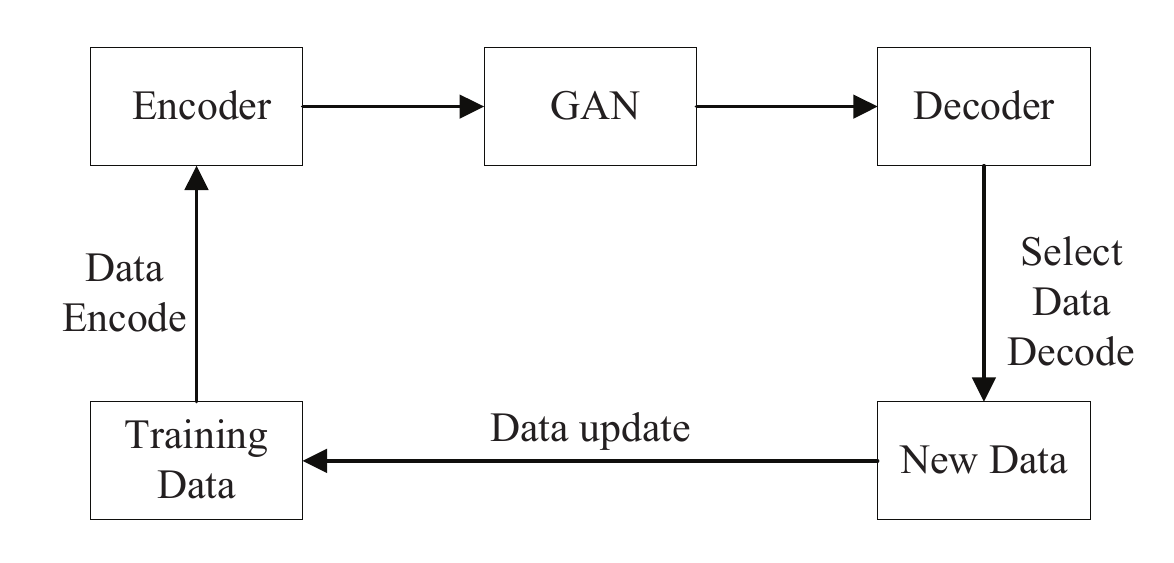}\\
  \caption{Iterative algorithmic framework of HpGAN. The training dataset is iteratively updated to improve the search capability of GAN until equilibrium is attained.}
\end{figure}

In what follows, we introduce these components and the relationship between them with more details.

\subsubsection{Encoder}
The encoder is designed not only to transform discrete data into continuous data, but also plays an important role in expanding the training dataset. The encoder is defined as follows:
\begin{Definition}
Let $\textbf{S}_{P}=\{\bm{x}_{1},\bm{x}_{2},\ldots,\bm{x}_{P}\}$ be a sample dataset with $P$ discrete sequences, then the corresponding continuous sample is
\begin{equation}\label{encode}
\bm{y}=\frac{1}{N}\left[\sum^{P}_{p=1}[\bm{x}_{p}(\bm{x}_{p})^{T}-\bm{I}]-\textbf{b}\right],
\end{equation}
where $N$ is the length of sequence $x_{p}$, $\bm{I}$ and $\bm{b}$ are the identity matrix and bias vector, respectively.
\end{Definition}
In HpGAN, in order to eliminate the dimensional influence between data features, we limit the data normalization to $[-1,1]$, i.e., $\frac{1}{N}\sum^{P}_{i=1}[\bm{x}_{p}(\bm{x}_{p})^{T}-\bm{I}]$. The bias vector $\bm{b}$ represents random noise, which is used to increase the tolerance of encoding. Another advantage is to improve the exploration ability of the model. The specific steps of the encoding process are as follows:
\begin{itemize}
             \item Choose a number $P$ less than $\mathcal{N}$ at random, i.e., $P\leq\mathcal{N}$, where $\mathcal{N}$ is a priori constant.
             \item Choose $P$ samples in $\mathcal{S}$ at random, and transform them into continuous sample by (\ref{encode}).
             \item Repeat the above steps until enough datasets are generated.
           \end{itemize}

In this way, we can obtain sufficient different datasets to effectively solve the problem of small samples. In the next subsection, we will explain the validity of continuous datasets.
\subsubsection{Decoder}
\begin{figure}
  \centering
  \includegraphics[width=3.0in]{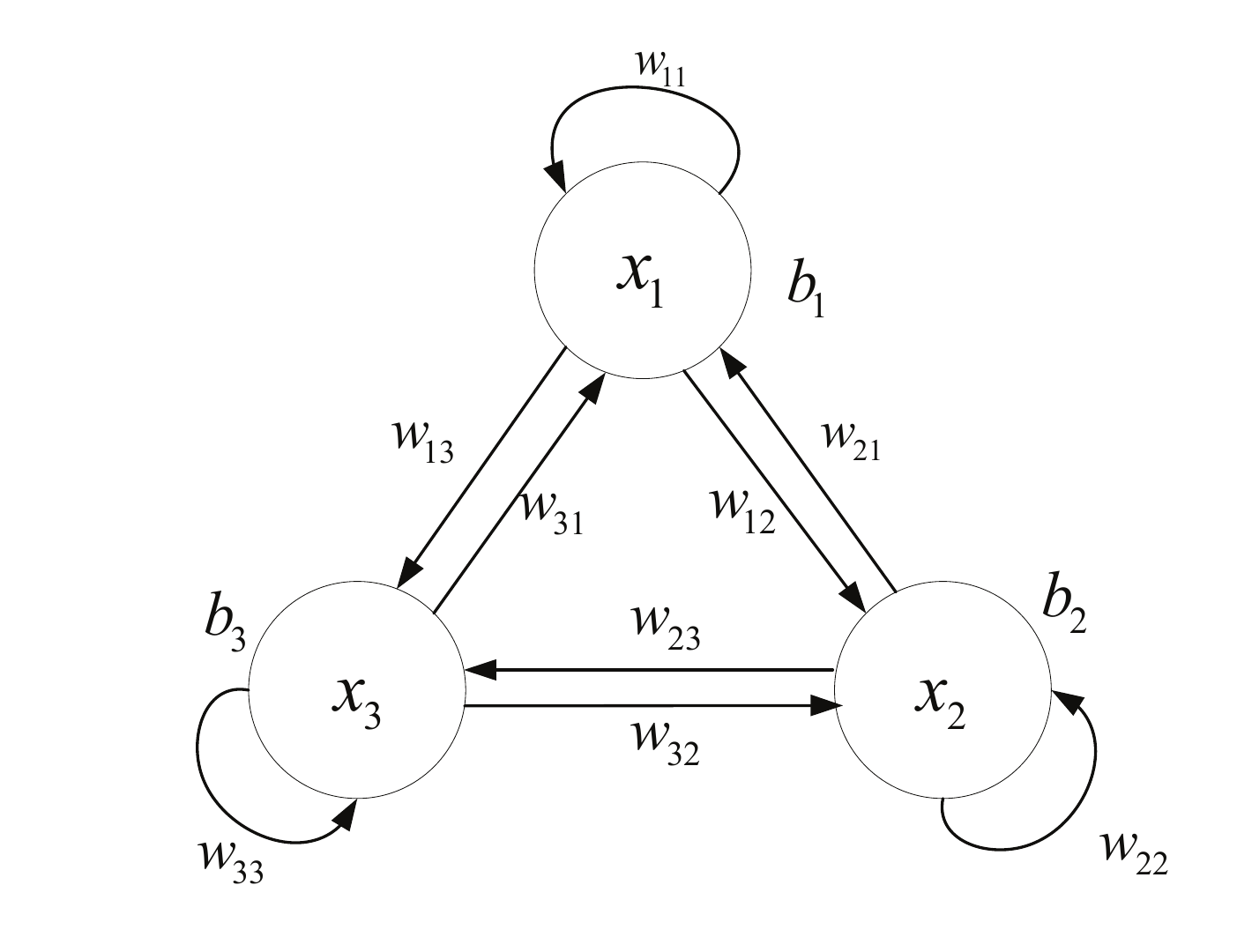}\\
  \caption{DHNN architecture with three neurons, in which neuron $i$ is connected to neuron $j$ with the weight $w_{ij}\in (-1,1)$, and $b_{i}$ is the bias of neuron $i$. Each neuron is connected to all neurons of the network.}
\end{figure}

In this subsection, we introduce our designed decoder based on discrete Hopfield neural network (DHNN) framework. DHNN is a single-layer full feedback neural network with $n$ neurons. Fig. 2 shows a DHNN architecture with three neurons, where $w_{ij}$ denotes the weight value connecting neuron $i$ to neuron $j$ and $b_i$ is the bias of neuron $i$. $x_i$  is called the state of neuron $i$, and the set $\bm{x}=\{x_{1},x_{2},x_{3}\}$ of all neuron states constitutes the state of feedback network. The input of the feedback network is the initial value of the network, which is expressed as $\bm{x}^{(0)}=\{x^{(0)}_{1},x^{(0)}_{2},x^{(0)}_{3}\}$. In the dynamic evolution process of the network from the initial state $\bm{x}^{(0)}$, the state transition rule of each neuron can be expressed as follows:
\begin{equation}\label{DHNN1}
  x_{j}=f(\mbox{net}_{j}), ~~~j=1,2,3,
\end{equation}
where $f(\cdot)$ denotes the activate function. Specifically, the symbolic function is used as $f(\cdot)$ in DHNN, i.e.,
\begin{equation}\label{DHNN2}
f(\mbox{net}_{j})=\mbox{sgn}(\mbox{net}_{j})=\left\{
                                 \begin{array}{cc}
                                   1, & \mbox{net}_{j}\geq0; \\
                                   -1, & \mbox{net}_{j}<0. \\
                                 \end{array}\right.
\end{equation}
The $\mbox{net}_i$ is the net input of neuron $i$, which be defined as
\begin{equation}\label{DHNN3}
\mbox{net}_{j}=\sum^{n}_{j=1}(w_{ij}x_{i}-b_{i}),  ~~~i=1,2,3,
\end{equation}
where $w_{ii}=0, w_{ij}=w_{ji}$.

In DHNN, there are two dynamic evolution methods: synchronous update and asynchronous update. In the synchronous update mode, all neurons in the network adjust the state at the same time. In contrast, in the asynchronous update mode, only one neuron state is adjusted while the others remain unchanged. In general, DHNN evolves faster in the synchronous update mode, but it is easy to fall into an infinite loop, and the opposite is true in asynchronous update mode.
\begin{Example}
Let us consider a DHNN with the same transition architecture illustrated in Fig. 2. The weight matrix and bias vector are $\bm{W}$, $\bm{b}$ respectively, where
\begin{eqnarray}\label{example1}
  \bm{W}&=&\left[
      \begin{array}{ccc}
        w_{11} & w_{21} & w_{31} \\
        w_{12} & w_{22} & w_{32} \\
        w_{13} & w_{23} & w_{33} \\
      \end{array}
    \right]=\left[
              \begin{array}{ccc}
                0 & -0.5 & 0.2 \\
                -0.5 & 0 & 0.6 \\
                0.2 & 0.6 & 0 \\
              \end{array}
            \right],\nonumber \\ \bm{b}&=&[b_{1},b_{2},b_{3}]=[-0.1,0,0.1].
\end{eqnarray}
Let the initial state of DHNN be $\bm{x}^{(0)}=\{x^{(0)}_{1},x^{(0)}_{2},x^{(0)}_{3}\}=\{1,-1,1\}$. Then, for the two update methods, the next state of the network is as follows:

In the synchronous update mode, the next state is $\bm{x}^{(1)}=\{x^{(1)}_{1},x^{(1)}_{2},x^{(1)}_{3}\}$, where
\begin{eqnarray}\label{example2}
  \left[
    \begin{array}{c}
      x^{(1)}_{1}\\
      x^{(1)}_{2}\\
      x^{(1)}_{3}\\
    \end{array}
  \right]=f\left(
            \textbf{W}\cdot
   \left[
    \begin{array}{c}
      x^{(0)}_{1}\\
      x^{(0)}_{2}\\
      x^{(0)}_{3}\\
    \end{array}
  \right]-  \left[
    \begin{array}{c}
      b_{1}\\
      b_{2}\\
      b_{3}\\
    \end{array}
  \right]\right)=  \left[
    \begin{array}{c}
      1\\
      1\\
      -1\\
    \end{array}
  \right].
\end{eqnarray}

In the asynchronous update mode, we randomly select a neuron to update its state, assuming $x^{(0)}_{1}$ is selected, then the next state $\bm{x}^{(1)}=\{x^{(1)}_{1},x^{(1)}_{2},x^{(1)}_{3}\}$, where
\begin{eqnarray}\label{example3}
  x^{(1)}_{1}&=&f\left(w_{11}\cdot x^{(0)}_{1}+w_{21}\cdot x^{(0)}_{1}+w_{31}\cdot x^{(0)}_{1}
-b_{1}
  \right)=-1,\nonumber\\
  x^{(1)}_{2}&=&x^{(0)}_{2},~x^{(1)}_{3}=x^{(0)}_{3}.
\end{eqnarray}
\end{Example}
In \cite{Hopfield1985}, the concept of energy function is introduced into the Hopfield neural network, which provides a reliable basis for judging the stability of network operation. The running process of DHNN is the evolution of states by dynamics. For an arbitrary initial state $\bm{x}^{(0)}$, it evolves in the way of energy reduction, and finally reaches a stable state $\bm{x}$. The stable state $\bm{x}$ is called the Hopfield network attractor which satisfies $\bm{x}=f(\bm{W}\bm{x}-\bm{b})$, where $\bm{W}$ and $\bm{b}$ are weight matrix and bias vector, respectively.
\begin{Theorem}
Let $\bm{X}_{P}=\{\bm{x}_{1},\bm{x}_{2},\ldots,\bm{x}_{P}\}$ be a sequence set, which  consists of $P$ binary sequences with same length $n$. Then in the synchronous update mode, the sequences $\bm{x}_{p}$ or $-\bm{x}_{p}$,$~p=1,2,\ldots,P$ which are the attractors of the DHNN with the weight matrix $\bm{W}=\sum^{P}_{p=1}[\bm{x}_{p}(\bm{x}_{p})^{T}-\textbf{I}]$ and no bias terms, i.e.,
\begin{small}
\begin{equation}\label{decode}
f(\bm{W} \bm{x}_{p})=\left\{
          \begin{array}{cc}
           \bm{x}_{p}, &  (n-P)-(m_{1}+m_{2}+\ldots+m_{P})\geq 0;\\
           -\bm{x}_{p}, & (n-P)-(m_{1}+m_{2}+\ldots+m_{P})<0,
           \end{array}\right.
\end{equation}
\end{small}
where $m_{k}=(x_{k})^{T}x_{p},~k=1,\ldots,p-1,p+1,\ldots,P$ and $f(\cdot)$ be the symbolic function.
\end{Theorem}
\textit{Proof:}
Since each symbol of the sequence set $\textbf{X}$ is $1$ or $-1$, we have
\begin{equation}\label{proof1}
(\bm{x}_{k})^{T}\bm{x}_{p}=\left\{
                                 \begin{array}{cc}
                                 m_{k}, &  p\neq k;\\
                                 n, & p=k,
                                 \end{array}\right.
~~~k=1,2,\ldots,n,
\end{equation}
where $-n\leq m_{p}\leq n$. Thus, we have
\begin{small}
\begin{equation}
\begin{aligned}
\bm{W} \bm{x}_{p}&=\sum^{P}_{k=1}[\bm{x}_{k}(\bm{x}_{k})^{T}-I]
\bm{x}_{p}=\sum^{P}_{k=1}[\bm{x}_{k}(\bm{x}_{k})^{T}\bm{x}_{p}-\bm{x}_{p}]\\
&=\bm{x}_{1}(\bm{x}_{1})^T\bm{x}_{p}+\ldots+\bm{x}_{p}(\bm{x}_{p})^T\bm{x}_{p}
+\ldots+\bm{x}_{P}(\bm{x}_{P})^T\bm{x}_{p}\\
&~~~-P\bm{x}_{p}\\
&=(m_{1}\bm{x}_{1}+m_{2}\bm{x}_{2}+\ldots+m_{P}\bm{x}_{P})+(n-P)\bm{x}_{p}.
\label{proof2}
\end{aligned}
\end{equation}
\end{small}
Since each symbol of sequences is $1$ or $-1$, (\ref{decode}) can be rewritten as
\begin{small}
\begin{equation}
\begin{aligned}
f(\bm{W} \bm{x}_{p})&=f[(m_{1}\bm{x}_{1}+m_{2}\bm{x}_{2}+\ldots+m_{P}\bm{x}_{P})+(n-P)\bm{x}_{p}]\\
&=\mbox{sgn}[(m_{1}\bm{x}_{1}+m_{2}\bm{x}_{2}+\ldots+m_{P}\bm{x}_{P})+(n-P)\bm{x}_{p}]\\
&=\left\{
          \begin{array}{cc}
           \bm{x}_{p}, &  (n-P)-(m_{1}+m_{2}+\ldots+m_{P})\geq 0;\\
           -\bm{x}_{p}, & (n-P)-(m_{1}+m_{2}+\ldots+m_{P})<0.
           \end{array}\right.
\label{proof3}
\end{aligned}
\end{equation}
\end{small}
This completes the proof.
$\hfill\blacksquare$
\begin{Remark}
In the asynchronous update mode, the equation in Theorem 1 is written as
\begin{small}
\begin{equation}\label{decode2}
f(\bm{W} \bm{x}_{p})=\left\{
          \begin{array}{cc}
           \bm{x}_{p}, &  (n-P)-(m_{1}+m_{2}+\ldots+m_{P})\geq 0;\\
           \bm{\hat{x}}, & (n-P)-(m_{1}+m_{2}+\ldots+m_{P})<0,
           \end{array}\right.
\end{equation}
\end{small}
where $\bm{\hat{x}}$ does not belong to the sequence set $\bm{X}_{P}$, meaning that DHNN converges to attractors other than $\bm{X}_{P}$. It is easy to see that the smaller of $P$, the greater the probability that $x_{p}$ is the attractor of DHNN. In this work, we selected a suitable $P$ through experiments and adopted a generalized asynchronous update mode. Specifically, DHNN randomly updates the state of $k$ neurons every time it evolves to avoid the drawbacks of synchronous and asynchronous update modes.
\end{Remark}

In HpGAN, DHNN is designed through the weight $\bm{W}$, which is  the continuous matrix that generated by the decoder with $P$ suitable binary sequences. It can be guaranteed that the decoder can successfully map to the original binary sequences when DHNN converges to a stable state by \textbf{Theorem 1}.
\begin{Example}
Consider a binary sequence set $\mathcal{C}=\{\bm{x}_{1},\bm{x}_{2},\cdots, \bm{x}_{M}\}$ which consists of $M$ different sequences of the same length $N=3$. Randomly select $P$ sequences from $\mathcal{C}$ and encode them as continuous data $\bm{W}$, where we assume $P=2, x_{1}=[1,1,-1],x_{2}=[1,-1,1]$. By \textbf{Definition 1} (no bias terms), we have
\begin{equation}\label{example2_1}
  \bm{W}=\left[
\begin{array}{ccc}
 0 & 0 & 0 \\
 0 & 0 & -\frac{2}{3} \\
  0 & -\frac{2}{3} & 0 \\
 \end{array}
 \right],
\end{equation}
then feed continuous data $\bm{W}$ as training data into GAN. When GAN reaches equilibrium, GAN generate a new continuous data $\bm{W}'$ which is similar to $\bm{W}$. We assume
\begin{equation}\label{example2_2}
  \bm{W}'=\left[
            \begin{array}{ccc}
              0.01 & 0 & 0 \\
              0 & 0 & -0.65 \\
              0 & -0.67 & 0 \\
            \end{array}
          \right]
\end{equation}
and exploit it as the weight matrix of DHNN with three neurons. Through the dynamic evolution of DHNN, we can know that for any initial state, when DHNN is stable, DHNN will always converge to $x_1$ or $x_2$.
\end{Example}
Overall, in one iteration, there are three main steps: 1) All two-dimensional continuous data generated by the encoder are used as training data of GAN; 2) GAN model is trained until it reaches Nash equilibrium; 3) The generator generates continuous two-dimensional data and converts it into binary sequences by the decoder. In the next iteration, we update the initial binary sequences with the generated better ones. In this way, the sequences generated by GAN gets improved after every iteration. Therefore, HpGAN has the ability to solve the problem of sequence search. The pseudocode for HpGAN is given in Algorithm 1.

\begin{flushleft}
\begin{tabular}{l}
  \hline
  \textbf{Algorithm 1}: HpGAN\\
  \hline
\textbf{Initialization:}\\
~~~Generate a binary sequence set  $\mathcal{S}$ as training data set\\
~~~through existing algorithms or constructions.\\
~~~Initialize the parameters $\textbf{W}$ and $\textbf{b}$ of GAN.\\
\textbf{while 1 do:}\\
1: Generate the continuous training dataset $\mathcal{S}_{con}$ by the\\
~~~ encoder.\\
2: Train GAN with the continuous dataset $\mathcal{S}_{con}$.\\
3: Generate continuous dataset $\mathcal{S}_{GAN}$ by the generator, \\
~~~until GAN reaches Nash equilibrium.\\
4: Transform $\mathcal{S}_{GAN}$ into binary sequences $\mathcal{S}_{bin}$ by the\\
~~~ decoder.\\
5: Compute metric $\mathcal{M(C)}$, and select $M$ sequences with\\
 ~~~good $\mathcal{M(C)}$ to constitute a new dataset $\mathcal{S}_{new}$, where \\ ~~~$\mathcal{C}\in\mathcal{S}_{bin}$.\\
6: Update dataset, i.e., $\mathcal{S}\leftarrow \mathcal{S}_{new}$.\\
\textbf{end}\\
\hline
\end{tabular}
\end{flushleft}

In the following sections, we demonstrate the searching capabilities of HpGAN in two applications: in Section III, we use HpGAN to search some MOCCSs which have the potential of enabling of interference-free MC-CDMA systems; in Section IV, we use HpGAN to search a new phase-coded sequence for pulse compression radar systems. In the two applications, in order to highlight the advantages of discrete sequences generated by HpGAN, we also compare the binary sequences generated by GAN and HpGAN, respectively.

\section{HpGAN for MOCCSs}
Let us consider a MOCCS $\mathcal{C}=\{\bm{c}_{j}^{m}(n): j=0,1,\ldots,J-1; m=0,1\ldots,M-1; n=0,1,\ldots,N-1\}$ which have zero auto- and cross- correlation sum properties for all non-trivial non-zero time-shifts. Specifically,
\begin{enumerate}
  \item \emph{Ideal Aperiodic Auto-correlation Function (AAF)}:
 For the $M$ sequences assigned to a user $j$, i.e., $\{c^{m}_{j}: m=0,1,\ldots,M-1\}$, the sum of the AAF of these sequences is zero for any nonzero shift:
\begin{equation}\label{AAF}
\hbox{AAF}_{j}(\tau)=\sum^{M-1}_{m=0}\sum^{N-1-\tau}_{n=0}c_{j}^{m}(n)c_{j}^{m}(n+\tau)=0,
\end{equation}
where delay $\tau=-N+1,\ldots,N-1, \tau\neq0$. Any sequence in this set is called a complementary set
sequence (CSS). In particular, when $M = 2$, the set is called
a Golay complementary pair (GCP), and any constituent sequence in this
pair is called a Golay sequence (GS) \cite{Tseng1972}.
\item Ideal Aperiodic Cross-correlation Function (ACF):
For two flocks of complementary codes assigned to users $j_{1}$ and $j_{2}$, i.e., $\{\bm{c}_{j_{1}}^{m},\bm{c}_{j_{2}}^{m}: m=0,1,\ldots,M-1\}$, the sum of their ACFs is always zero irrespective of the relative shift:
\begin{equation}\label{ACF}
\hbox{ACF}_{j_{1},j_{2}}(\tau)=\sum^{M-1}_{m=0}\sum^{N-1-\tau}_{n=0}c_{j_{1}}^{m}(n)c_{j_{2}}^{m}(n+\tau)=0,
\end{equation}
where $\tau=-N+1,\ldots,N-1$ and $j_{1}\neq j_{2}$.
\end{enumerate}

Some known constructions of MOCCS are available in \cite{Chen2007}. In the next subsection, we make use of HpGAN to search MOCCS. Our goal is to investigate and evaluate the capability of HpGAN, i.e., whether it can search some  MOCCSs which are not in the training dataset.
\subsection{HpGAN for MOCCSs}
In this subsection, we use HpGAN to search MOCCSs. As mentioned above, a MOCCS should satisfy equations (\ref{AAF}) and (\ref{ACF}) at the same time. Hence, we consider the following metric function $\mathcal{M(C)}$ which is the sum of all the non-trivial aperiodic auto- and cross- correlation squares of a sequence set $C$:
\subsubsection{Metric Function}
For a binary sequence set $\mathcal{C}=\{\bm{x}^{m}_{j}(n): j=0,1,\ldots,J-1; m=1,2,\ldots,M-1; n=0,1,\ldots,N-1; x^{m}_{j}(n)\in \{-1,1\}\}$ consisting of $MJ$ sequences of same length $N$, the metric function $\mathcal{M(C)}$ is defined below:
\begin{equation}
\begin{aligned}
  \mathcal{M(C)}&=\sum^{J-1}_{j=0}\sum^{N-1}_{\tau=-N+1}|\hbox{AAF}_{j}(\tau)|\\
  &~~~~~~+\sum^{J-1}_{j_1=0}\sum^{J-1}_{j_2=j_1+1}
  \sum^{N-1}_{\tau=-N+1}|\hbox{ACF}_{j_1,j_2}(\tau)|.\\
  \label{metric}
\end{aligned}
\end{equation}
For MOCCSs, it is desired to have $\mathcal{M}^{*}=\hbox{inf}~\mathcal{M(C)}=0$. Our goal is to demonstrate the effectiveness of HpGAN by generating new MOCCSs.
\subsubsection{Training Dataset}
Generating sufficient training data is a prerequisite of using HpGAN, and the quality of the training sequence sets determines the performance of HpGAN. In this paper, we exploit some known constructions in [12] to generate a training datasets $\mathcal{S}_{train}$, which contains 200 different MOCCSs, where $\mathcal{M}(\mathcal{C})=0, ~\mathcal{C}\in \mathcal{S}_{train}$ and $J=2, M=2, N=8$ for each $\mathcal{C}$.
\subsubsection{Encoder}
There are two main considerations for setting up an encoder: one is to convert discrete data into continuous data, and the other is to increase the amount of training data. We regard each sequence set $\mathcal{C}$ as a sequence of length 32. The encoder is designed by Definition 1. We generated 500 continuous data as the training data set of GAN by (\ref{encode}) with random $P$ and $\textbf{b}$. In general, the larger of $P$ and $\textbf{b}$, the more samples can be generated, but the lower of the decoding accuracy. From Fig. 3, we can see that there is a decoding accuracy when both $P$ and $\textbf{b}$ take relatively small numbers. Therefore, considering the diversity of generated data and the decoding accuracy, we set $P\leq 4$ and $\textbf{b}\in [0,0.4]$ in HpGAN.
\begin{figure}
  \subfigure[Decoding accuracy with the change of parameter $P$.]{
    \begin{minipage}{8cm}
      \centering
      \includegraphics[width=3.0in]{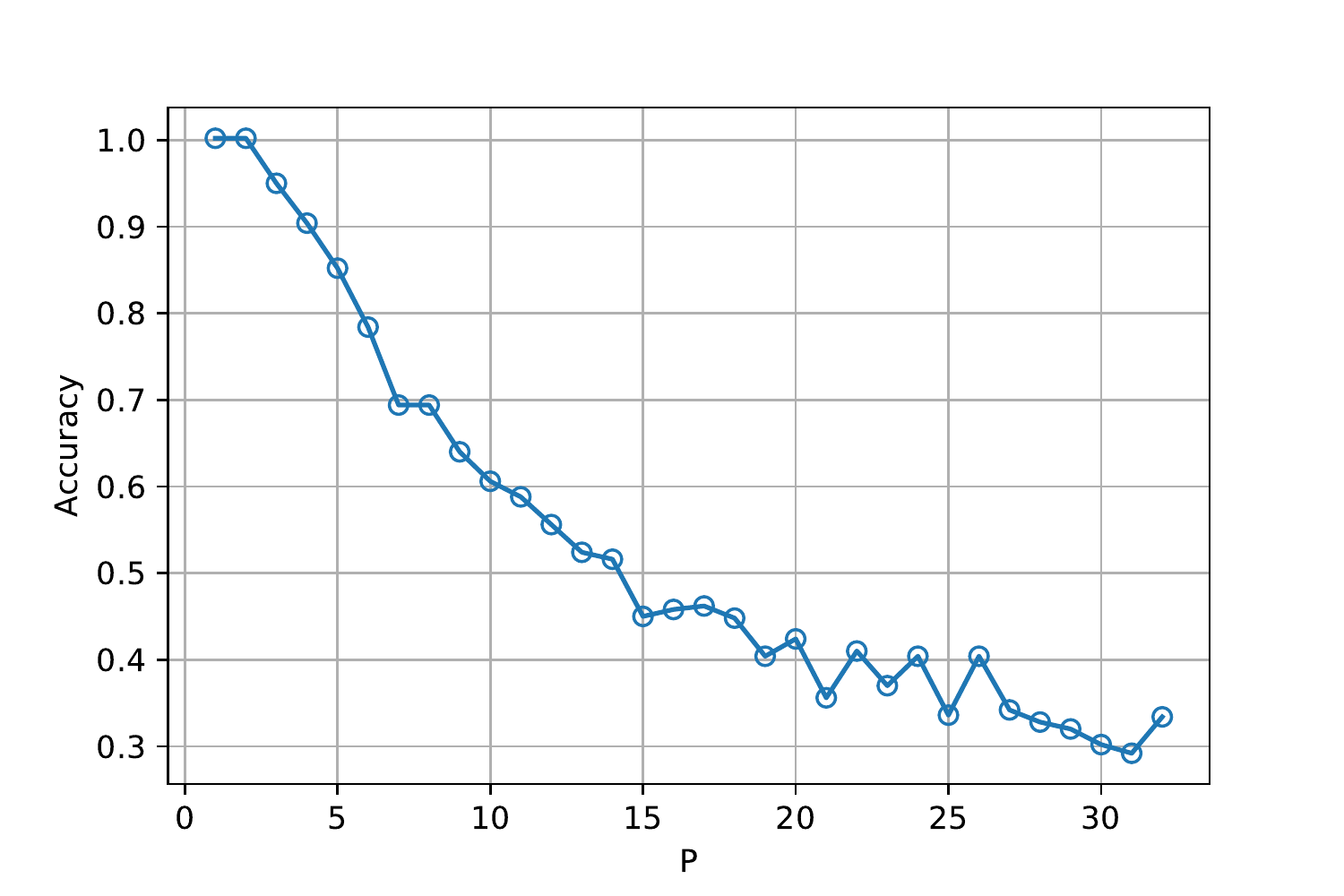}
    \end{minipage}}%

  \subfigure[Decoding accuracy with the change of parameter $\textbf{b}$]{
    \begin{minipage}{8cm}
      \centering
      \includegraphics[width=3.0in]{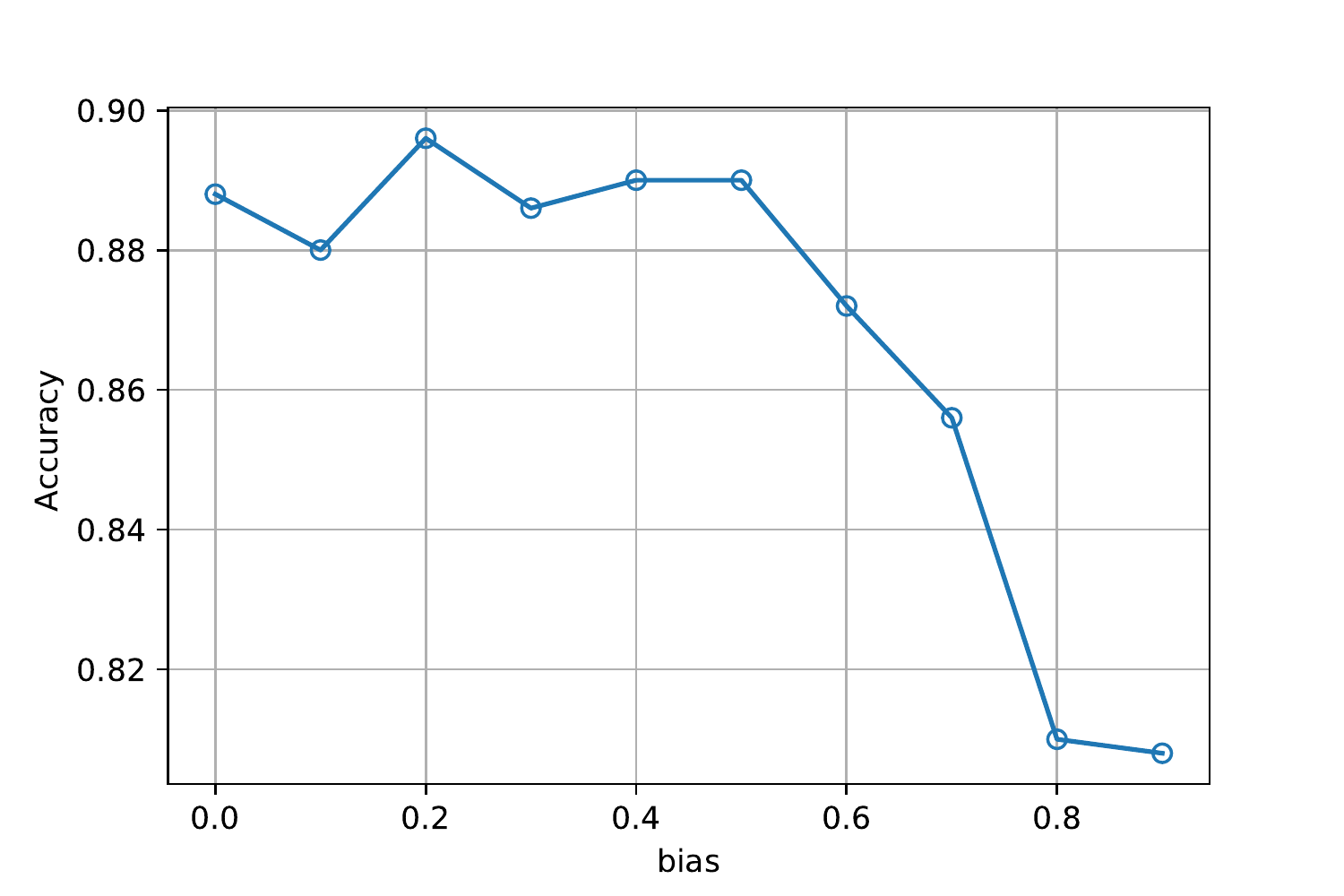}
    \end{minipage}}
  \caption{Illustration of the decoding accuracy with the two parameters $P$ and bias.}
\end{figure}

\subsubsection{GAN}
Both the generative model $G$ and the discriminant model $D$ in HpGAN are implemented by multilayer perceptrons. Each perceptron contains three layers of neurons: input layer, hidden layer and output layer. The input of the generator is a random noise vector $\textbf{z}$, where $\textbf{z}$ obeys uniform distribution, i.e., $\textbf{z}\sim U(-1,1)$.  In the process of training GAN, the minibatch size was set to 100, and we randomly sampled 5 minibatches without replacement from the 500 continuous data to train the GAN. The parameter settings of HpGAN are listed in Table I. $O_{G}$ and $I_{D}$ were set to 1024. Since the perceptron model can only process one-dimensional data, we converted all training data into one-dimensional data with a length of $32\times32$ and fed them into HpGAN. The discriminator is equivalent to a two-class model, thus $O_{D}$ was set to 1. Other parameters may be modified according to the effectiveness of the model.
\begin{table*}[ht!]
\caption{The parameters of HpGAN}
\centering
\begin{tabular}{c|c|c}
  \hline
  \textbf{Items} & \textbf{Parameters} & \textbf{Definition} \\
  \hline
  \multirow{3}*{Generative Model} & $I_{G}=100$ & Number of neurons in the input layer \\
  \cline{2-3}
    &$H_{G}=1024$ & Number of neurons in the hidden layer \\
  \cline{2-3}
    &$O_{G}=1024$ & Number of neurons in the output layer \\
  \hline
  \multirow{3}*{Discriminant Model} & $I_{D}=1024$ & Number of neurons in the input layer \\
  \cline{2-3}
    &$H_{D}=1024$ & Number of neurons in the hidden layer \\
  \cline{2-3}
    &$O_{D}=1$ & Number of neurons in the output layer \\
  \hline
  \multirow{2}*{Generative Model/Discriminant Model}
    &$\alpha=0.0001$ & Learning rate of the generative/discriminant model \\
  \cline{2-3}
    & $\hbox{batch}=100$ & Mini-batch size \\
  \hline
\end{tabular}
\end{table*}

\subsubsection{Decoder}
The decoder is a DHNN with 32 neurons which uses the generated data as its weight matrix, and then decodes the generated data through the dynamic evolution of DHNN. We employ asynchronous updates to evolve DHNN. Specifically, during each update of DHNN, only five neurons are randomly selected to update. For example, when only one neuron $i$ is updated, the evolution formula of DHNN is as follows:
\begin{equation}\label{decoder-eq1}
  x_{j}(t+1)=\left\{
               \begin{array}{ll}
                 \hbox{sgn}(\hbox{net}_{j}(t)), & j=i; \\
                 x_{j}(t), & j\neq i,
               \end{array}
             \right.
~~~j=1,2,\ldots,n.
\end{equation}
By this method, the search space can be increased, making it easier to get better results.
\subsection{Performance Evaluation}
We evaluated the search performances of HpGAN on a PC with Intel(R) Xeon(R) W-2125 CPU@ 4Ghz and 16GB RAM. In this experiment, the number of iterations of the network was set to $ 10^4$. In each iteration, we calculated the loss values of the generative model and the discriminant model to observe the convergence of HpGAN. To monitor the evolution of HpGAN, we evaluated the search ability of HpGAN every 100 iterations by calculating the metric $\mathcal{M(C)}$ of 100 sequence sets generated by the current generator. We exploited the minimum metric $\hbox{min}[\mathcal{M}]$ and the mean metric $E[\mathcal{M}]$ in the generated sequence sets to reflect the search effect and search trend of HpGAN.

The loss values of the generator and discriminator of HpGAN are shown Fig. 4. From Fig .4, we can see that after 2500 iterations, starts to converge, where the loss values of both the generative model and the discriminant model become stable. The $E[\mathcal{M}]$ and the minimum metric $\hbox{min}[\mathcal{M}]$ of the 100 generated sequence sets were recorded and plotted in Fig. 5. As seen from Fig. 5, after 2500 iterations, the value of $\hbox{min}[\mathcal{M}]$ reaches the optimal value, whilst $E[\mathcal{M}]$ exhibits a downward trend Fig. 4 and Fig. 5 show that HpGAN is effective in the search of complementary sequences. Indeed the generator has learned the characteristics of the training sequence sets, when HpGAN converges to a stable state.

We compare the performance of HpGAN and GAN in sequence generation tasks to demonstrate the necessity of encoding in HpGAN. We used the MOCCSs as training data and fed it to GAN without encoding, where GAN and HpGAN have the same structure. It can be seen from Fig. 6 that GAN fails to attain equilibrium, because the loss values of HpGAN have not converged, especially the loss values of the generator. From Fig. 7, we can see that in the GAN evolution process, the minimum metric $\hbox{min}[\mathcal{M}]$ and the mean metric $E[\mathcal{M}]$ of the sequence sets generated by the generator are not significantly improved. This shows that it is difficult for GAN to learn the effective features of the training sequences, when GAN is used to directly train binary sequences without encoding.

In this experiment, we found 119 different MOCSSs with $\mathcal{M(C)}=0$ by HpGAN, and none of these sequence sets belong to the training sequence sets. It is worth noting that the purpose of HpGAN is to train a generator so that it can generate data with similar characteristics to the training data. Once the training is completed, we can directly use the generator to quickly generate a large amount of new data without retraining.

\begin{figure}[h]
  \subfigure[The loss values of the discriminator with the change of the number of iterations]{
    \begin{minipage}{8cm}
      \centering
      \includegraphics[width=3.0in]{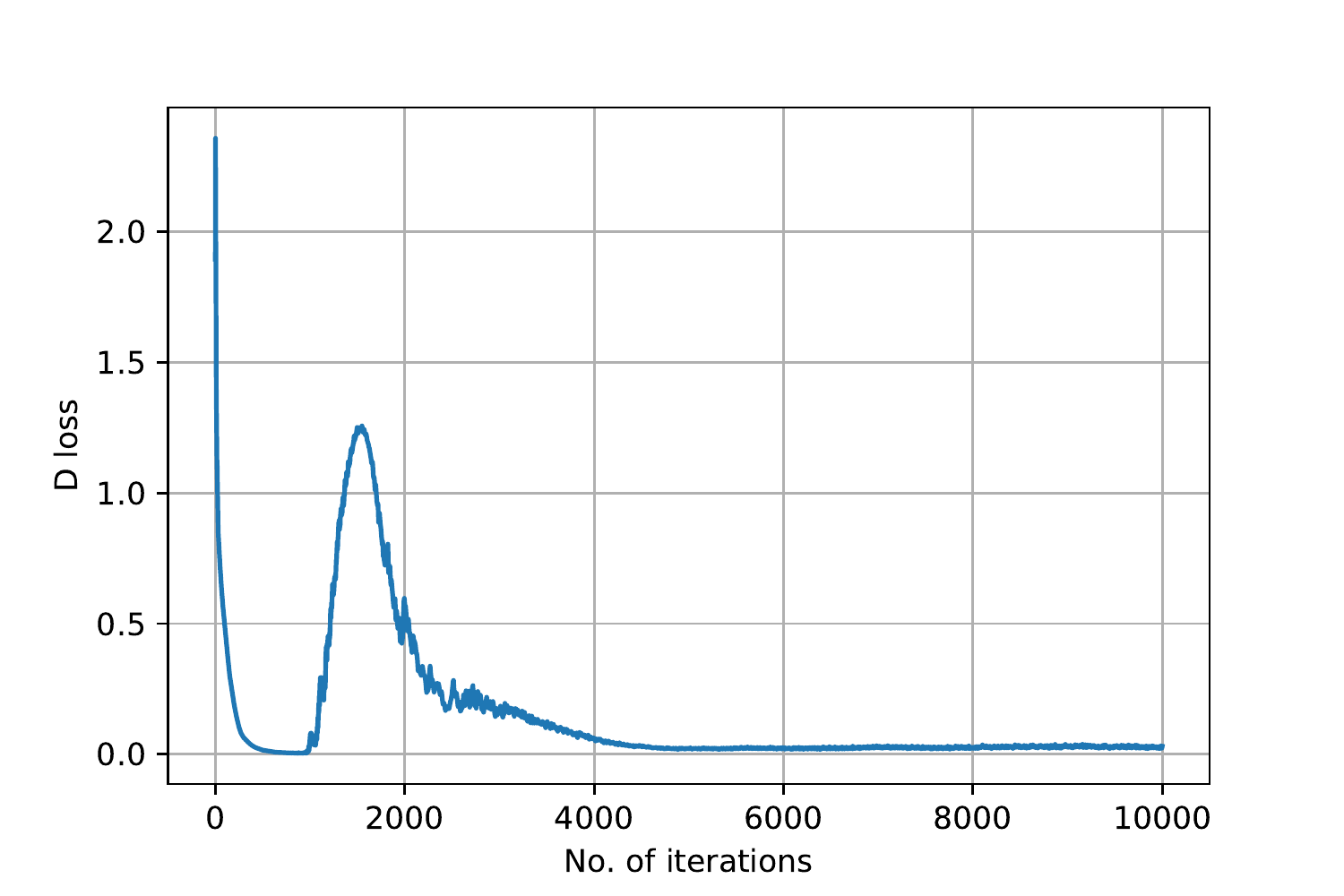}
    \end{minipage}}%

  \subfigure[The loss values of the generator with the change of the number of iterations]{
    \begin{minipage}{8cm}
      \centering
      \includegraphics[width=3.0in]{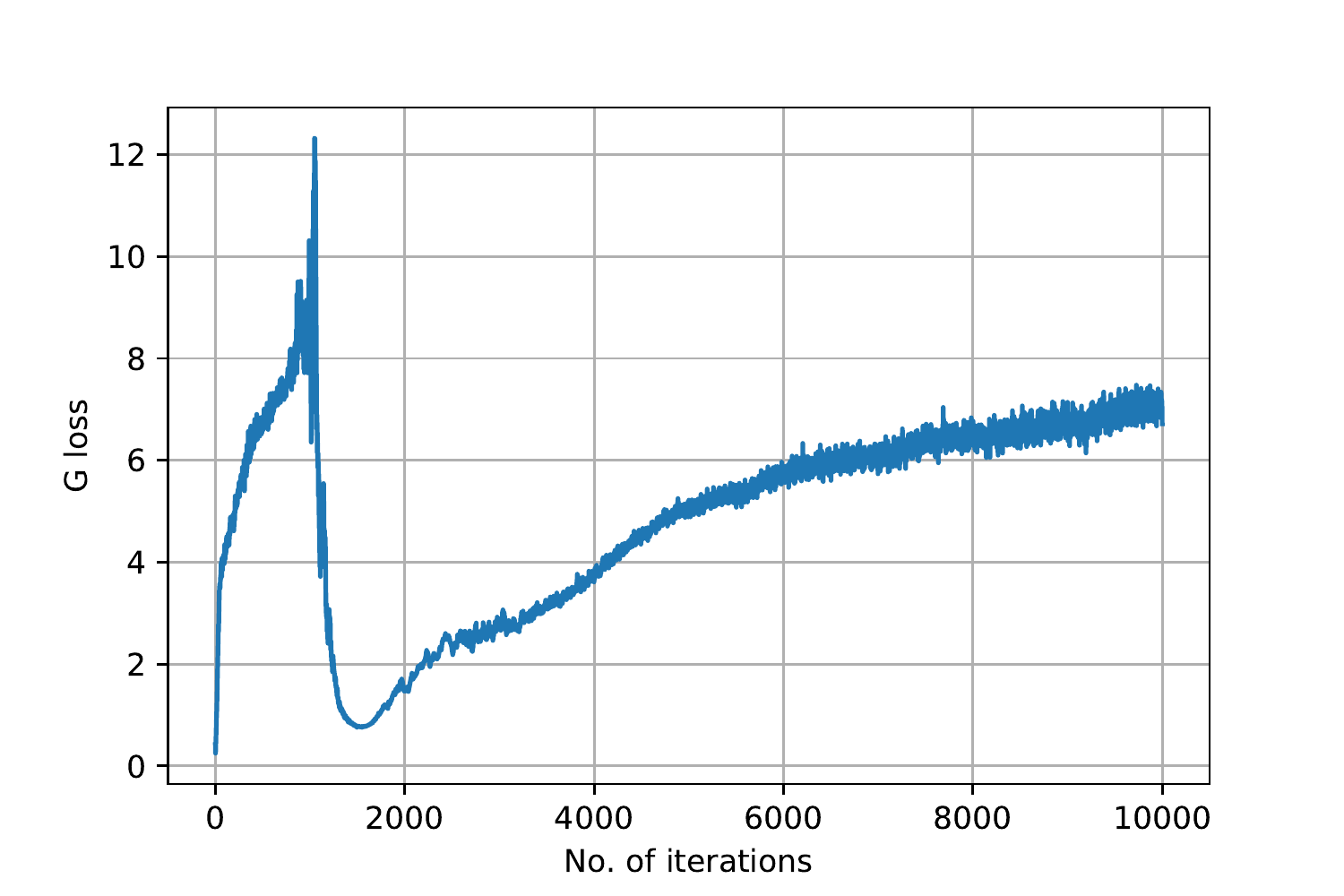}
    \end{minipage}}
  \caption{ The loss values of the generator and the discriminator
of HpGAN with the change of the number of iterations. After 2500 iterations, the values of the generative model and the discriminant model tend to be stable.}
\end{figure}
\begin{figure}[h]
  \centering
  \includegraphics[width=3.0in]{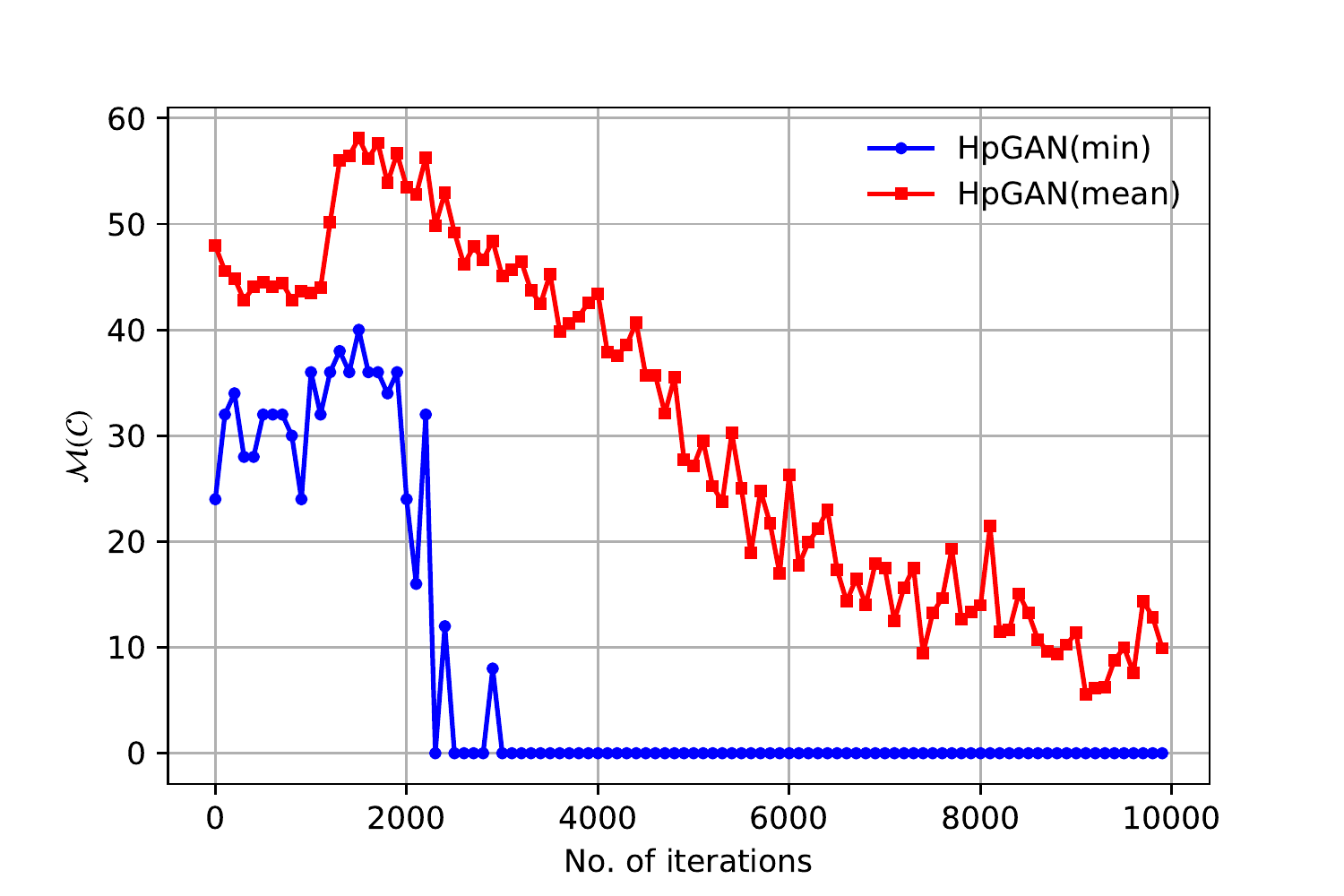}\\
  \caption{A training process of HpGAN to search a set of complementary sequences with the evolution curves of the mean metric $E(\mathcal{M})$ and the minimum metric $\hbox{min}[\mathcal{M}]$. After 2500 iterations, the minimum metric $Min[\mathcal{M}]$ reaches the optimal value, i.e., $Min[\mathcal{M}]=0$, and the mean metric $E[\mathcal{M}]$ gradually decreases.}
\end{figure}

\begin{figure}[h]
  \subfigure[The loss values of the discriminator with the change of the number of iterations]{
    \begin{minipage}{8cm}
      \centering
      \includegraphics[width=3.0in]{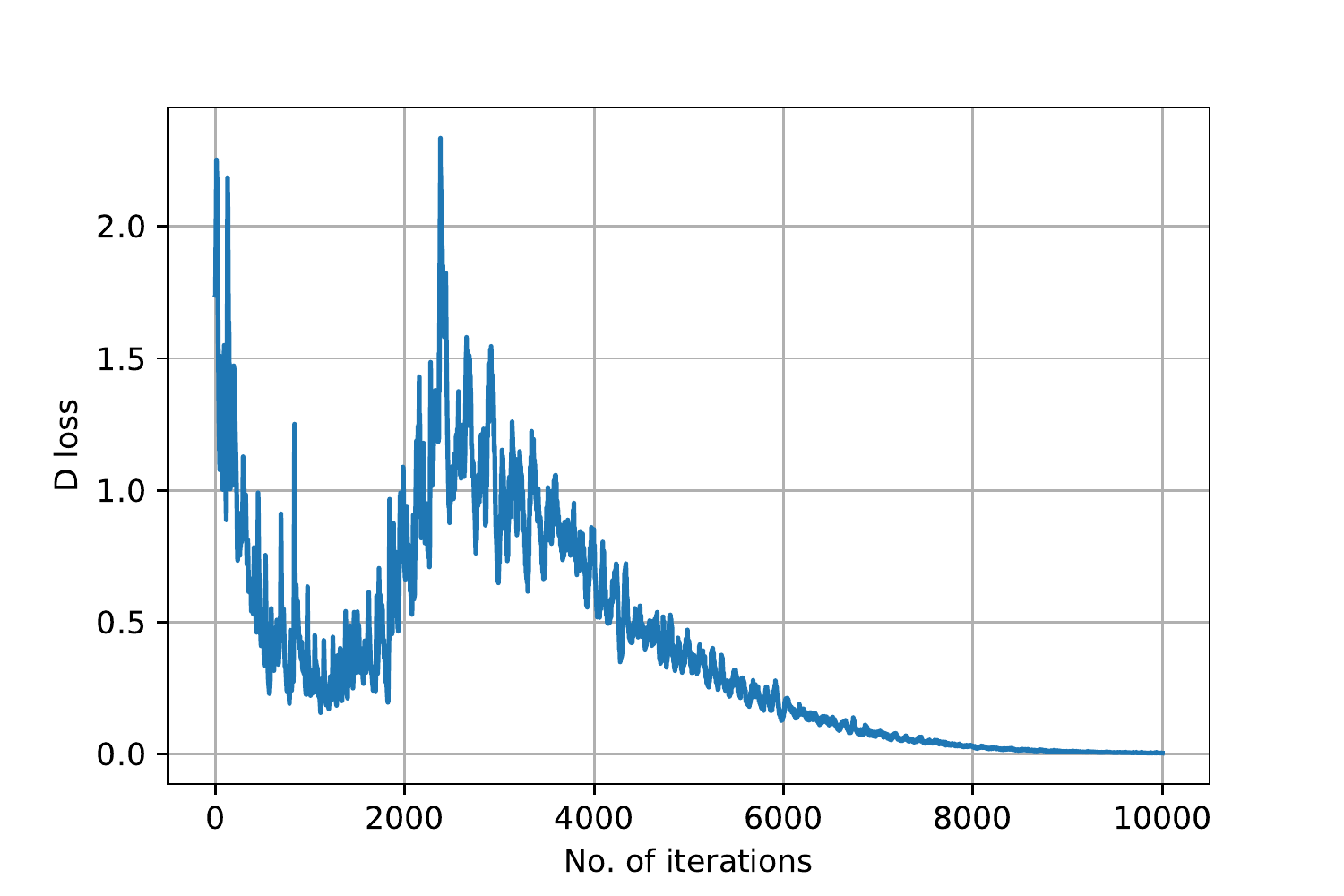}
    \end{minipage}}%

  \subfigure[The loss values of the generator with the change of the number of iterations]{
    \begin{minipage}{8cm}
      \centering
      \includegraphics[width=3.0in]{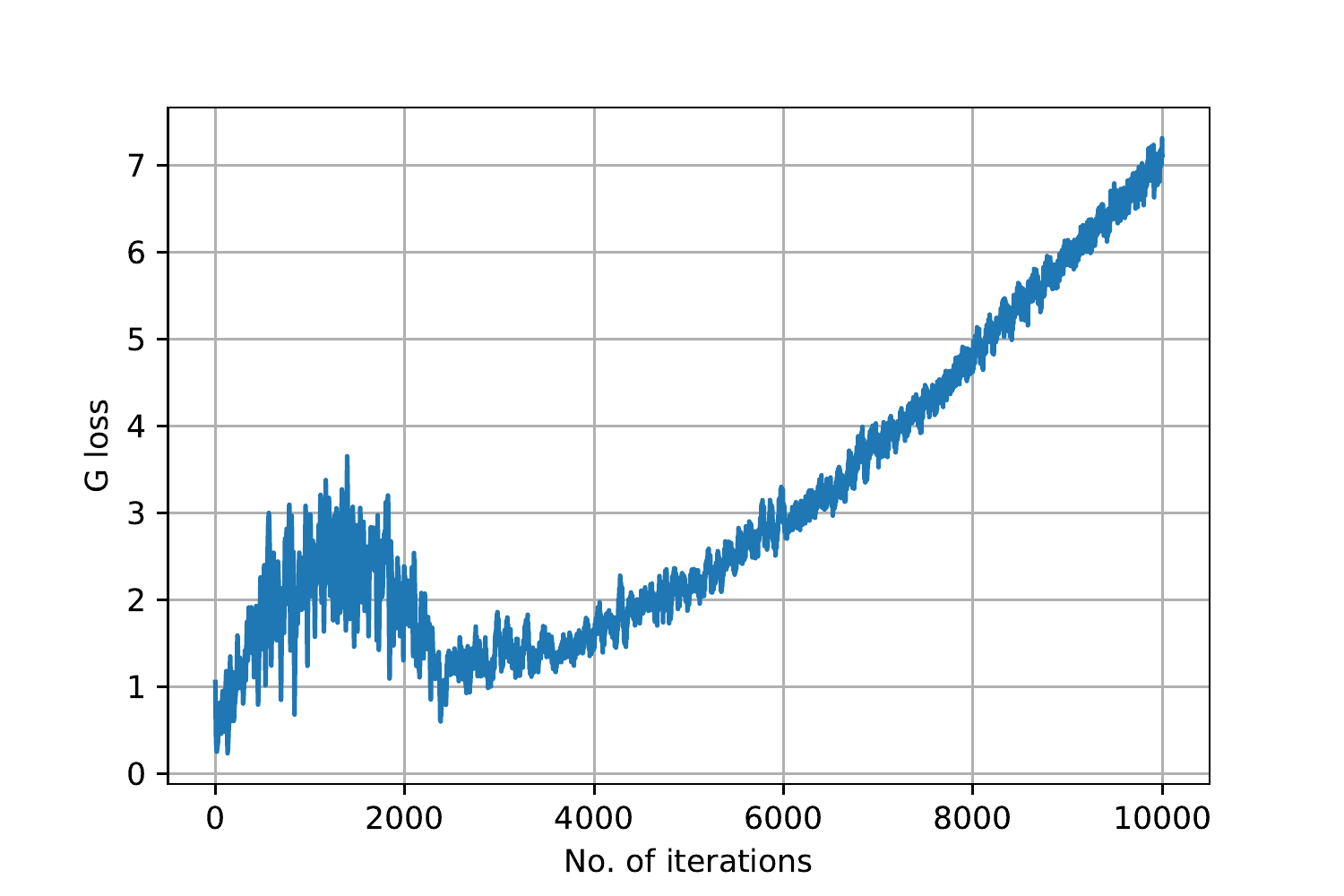}
    \end{minipage}}
  \caption{ The loss values of the generator and discriminator
of GAN. During the training process, the loss values of the discriminator gradually approaches zero, while the loss values of the generator shows a rapid increase.}
\end{figure}
\begin{figure}[h]
  \centering
  \includegraphics[width=3.0in]{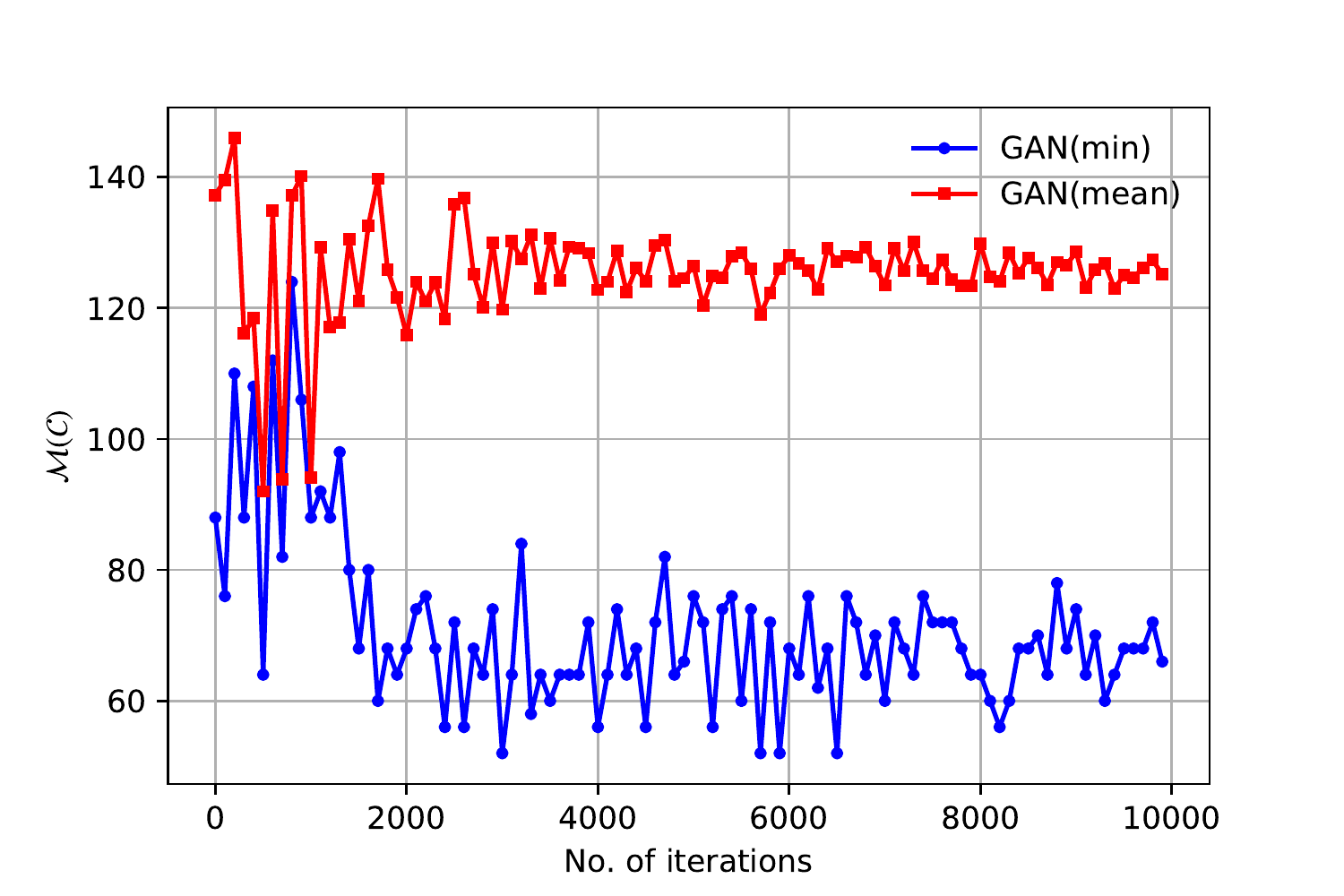}\\
  \caption{Training process of GAN to search a set of complementary sequences with the evolution curves of the mean metric $E(\mathcal{M})$ and the minimum metric $\hbox{min}[\mathcal{M}]$. During the training process, no significant improvements of $E(\mathcal{M})$ and $\hbox{min}[\mathcal{M}]$ have been attained.}
\end{figure}

\subsection{HpGAN for Optimal and Sub-Optimal OB-ZCPs}
In this subsection, to further demonstrate that the sequences generated by HpGAN may not be produced by systematic construction, we search optimal and sub-optimal OB-ZCPs by HpGAN.
In this experiment, we constructed 128 of OB-ZCPs of length 15 by the approach  in \cite{Liu2014} as the training data. This was done by deleting the first or last element of the binary GCPs of length 16. In the training data, the maximum ZCZ length of all OB-ZCPs is 4. In addition, the minimum PMEPR (interested readers may refer to \cite{Liu2014,Popovic1991,Davis1999} and references therein for more details) of each sequence in the training data is 1.7272. We designed HpGAN whose network structure and parameters are the same as those of HpGAN when searching for MOCCSs in the previous section.

After the experiment, we found 140 OB-ZCPs, where none of them belong to the training data. In particular, an optimal  Type-I OB-ZCP $s_\textrm{Type-I}$, Type-II OB-ZCP $s_\textrm{Type-II}$ and sub-optimal OB-ZCP $s_\textrm{sub-optimal}$ found by HbGAN are given as follows:
\begin{small}
\begin{eqnarray*}
\setlength{\arraycolsep}{0.1pt}
  s_\textrm{Type-I} =\left[
                         \begin{array}{ccccccccccccccc}
                            -1& -1&+1&-1&-1&+1 &-1&-1&-1&+1&-1&+1&-1&+1&+1 \\
                            -1& +1&+1& +1& +1&-1&-1&-1& +1&+1&-1&+1& +1&+1&+1\\
                         \end{array}
                       \right],
   \\
  \setlength{\arraycolsep}{0.1pt}
  s_\textrm{Type-II} =\left[
                         \begin{array}{ccccccccccccccc}
                            -1&+1&-1&+1&-1&-1&-1&-1&+1&+1&-1&+1&+1&-1&-1 \\
                            +1& +1& +1& +1& +1&-1& +1& +1& +1&-1&-1&-1& +1& +1&-1\\
                         \end{array}
                       \right],
  \\
\setlength{\arraycolsep}{0.1pt}
  s_\textrm{sub-optimal}=\left[
                         \begin{array}{ccccccccccccccc}
                            -1 &+1 &+1&-1&-1 &+1 &+1 &+1&-1&-1&-1&-1&-1 &+1&-1 \\
                            -1&-1&+1&-1&+1&-1&-1&-1&-1&-1&-1&+1&+1&-1&+1\\
                         \end{array}
                       \right].
\end{eqnarray*}
\end{small}

\begin{figure}
  \subfigure[For Type-I OB-ZCP $s_\textrm{Type-I}$.]{
    \begin{minipage}{8cm}
      \centering
      \includegraphics[width=3.0in]{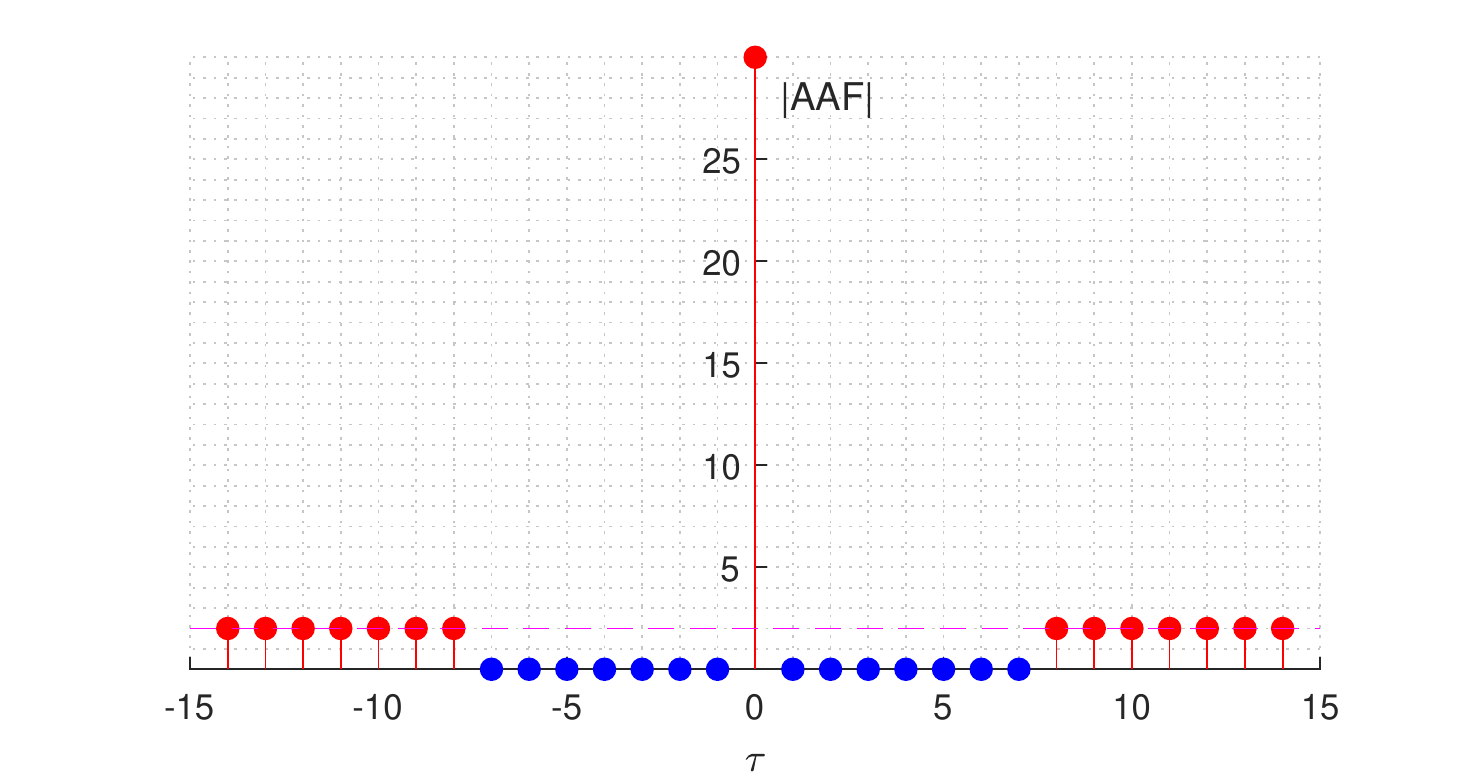}
    \end{minipage}}%

  \subfigure[For Type-II OB-ZCP $s_\textrm{Type-II}$.]{
    \begin{minipage}{8cm}
      \centering
      \includegraphics[width=3.0in]{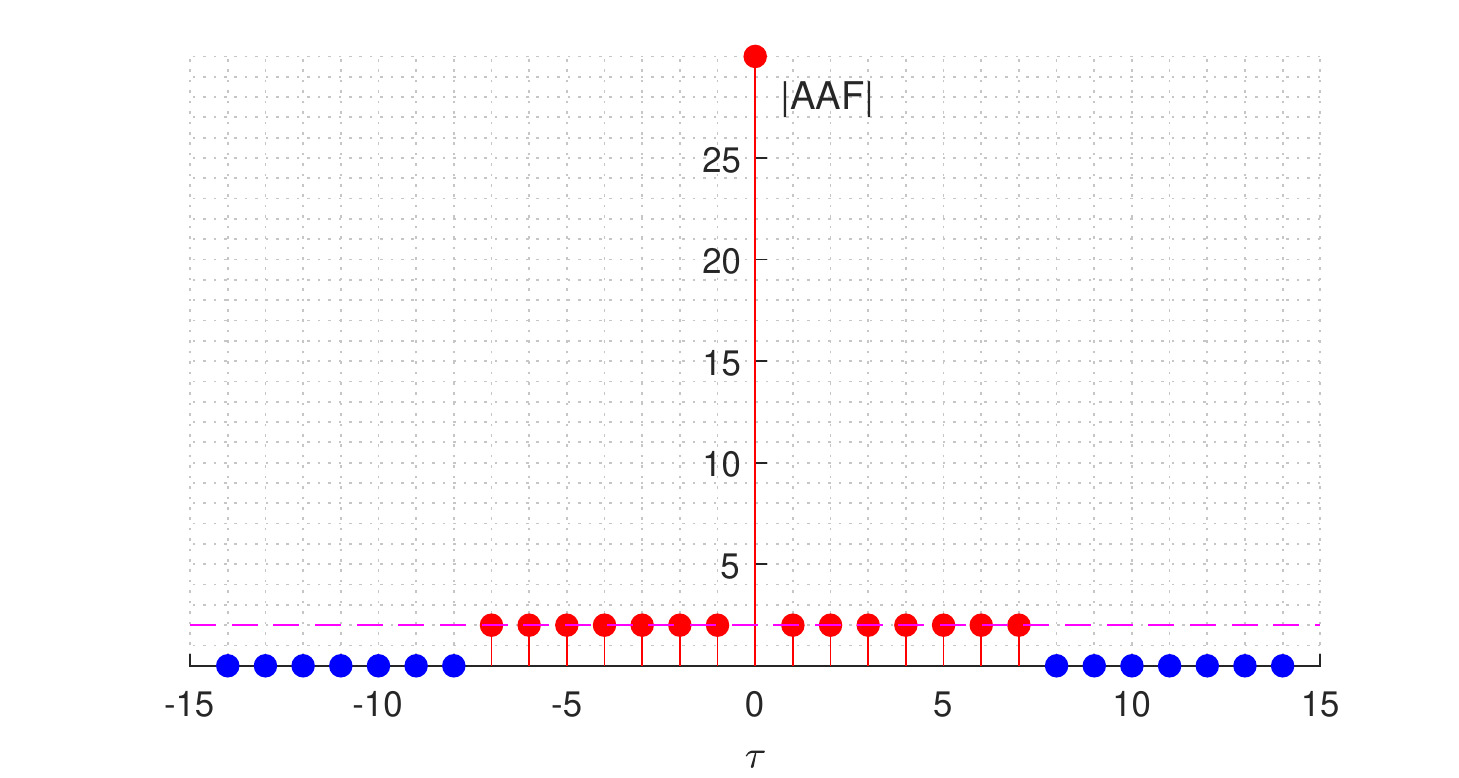}
    \end{minipage}}

  \subfigure[For sub-optimal OB-ZCP $s_\textrm{sub-optimal}$.]{
    \begin{minipage}{8cm}
      \centering
    \includegraphics[width=3.0in]{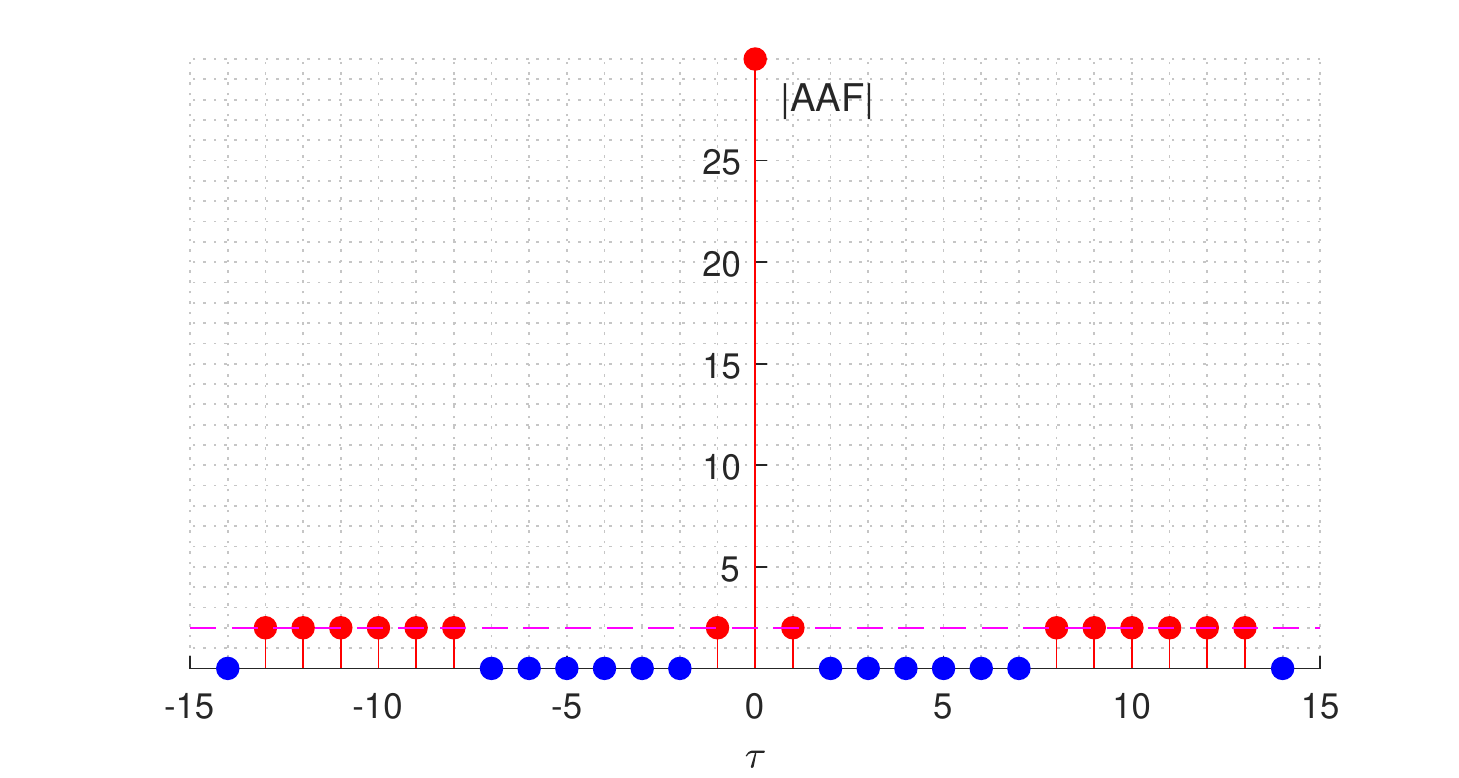}
  \end{minipage}}
  \caption{AAF sum magnitudes of OB-ZCP $s_\textrm{Type-I}$, OB-ZCP $s_\textrm{Type-II}$ and sub-optimal OB-ZCP $s_\textrm{sub-optimal}$, respectively.}
\end{figure}
A plot of their individual aperiodic auto-correlation sum magnitudes is shown in Fig. 8. One can see that the $s_\textrm{Type-I}$ and $s_\textrm{Type-II}$ are optimal. An sub-optimal OB-ZCP $s_\textrm{sub-optimal}$ found by HpGAN is shown in Fig. 9.
HpGAN also found the sequence with PMEPR metric $\textrm{PMEPR}(s_{\textrm{PMEPR}})\approx 1.6667$
\begin{small}
\begin{equation*}
\setlength{\arraycolsep}{0.1pt}
  s_{\textrm{PMEPR}}=\left[
                         \begin{array}{ccccccccccccccc}
                             -1 &-1 &-1 & +1 & +1 & +1 &-1 & +1 &-1 & +1 &-1 &-1 & +1 &-1 &-1 \\
                         \end{array}
                       \right],
\end{equation*}
\end{small}
which is better than any sequence in the training data.

\section{HpGAN for Pulse Compression Radar}

Let $\bm s$ be a binary probing sequence of length $N$, and $\bm y$ denotes the received sequence of length $N$. Denote by $\{h_{n}\}^{N-1}_{n=-N+1,n\neq 0}$ denote the corresponding amplitude coefficients for the adjacent range bins or clutter patches. $\textbf{J}_{n}$ denote a shift matrix that takes into account the fact which the clutter returns from adjacent range bins need different propagation times to reach the radar receiver \cite{Davis2007}:
\begin{equation}\label{radar1}
\setlength{\arraycolsep}{0.5pt}
 \textbf{J}_{n}=\left[
   \begin{array}{lccc}
     \overbrace{0~~~~\cdots~~~~0}^{n}&1&~&\textbf{0} \\
     ~ & ~ &\ddots& ~\\
     ~&~&~& 1\\
    \textbf{0} & ~ & ~ & ~ \\
   \end{array}
 \right]_{N\times N}=\textbf{J}^{T}_{-n}, ~~~n=1,2,\ldots,N-1.
\end{equation}
Following the definition in \cite{Davis2007} and \cite{Stoica2007}, after subpulse-matched filtering (MF) and analog-to-digital conversion, we can write the received sequence $\bm y$ as:
\begin{equation}\label{radar2}
  \bm{y}=h_{0}\bm{s}+\sum^{N-1}_{n=1-N,n\neq{0}} h_{n}\textbf{J}_{n}\bm{s}+\bm{w},
\end{equation}
where $\bm{w}$ denotes the additive white Gaussian noise (AWGN).

Given the received sequence $\bm{y}$, the radar's objective is to estimate $h_{0}$, where $h_{0}$ corresponds to the range bin of interest. To this end, the operation of a conventional receiver is described by the following equation \cite{Davis2007, Golay1983, Stoica2007, Golay1982, Hoholdt2006, Jedwab2008}:
\begin{equation}\label{estimator}
  \hat{h}_{0}=\frac{\bm{s}^{T}\bm{y}}{\bm{s}^{T}\bm{s}}=h_{0}+
  \sum^{N-1}_{n=1-N,n\neq0}h_{n}\frac{\bm{s}^{T}\textbf{J}_{n}\bm{s}}
  {\bm{s}^{T}\bm{s}},
\end{equation}
where the additive white Gaussian noise is ignored since the received signal is interference-limited. Because $h_{0}$ and $\{h_{n}\}^{N-1}_{n=-NN+1,n\neq 0}$ are unknown, it is reasonable to define the signal-to-interference ratio (SIR) $\gamma_{\textrm{MF}}$ at the output of the above receiver is as follows:
\begin{equation}\label{MF}
 \gamma_{\textrm{MF}}=\frac{(\bm{s}^{T}\bm{s})^2}{\sum^{N-1}
 _{n=1-N,n\neq0}(\bm{s}^T\textbf{J}_{n}\bm{s})^2}.
\end{equation}

There is a rich body of literature on maximizing $\gamma_{\textrm{MF}}$ over the set of binary sequences, which is referred to as the well-known ``merit factor problem'' \cite{Golay1982, Hoholdt2006, Jedwab2008, Jedwab2005, Golay1977, Ferguson2005}.
For such sequences, the best-known merit factor of 14.08 is achieved by the Barker sequence of length 13 \cite{Barker1953}.

In the presence of Gaussian white noise, the above MF estimator can provide the largest signal-to-noise ratio (SNR). However, some clutters may cause interference to the received information, especially in the scene of a weak target detection. Hence, interference suppression is important. This motivates the design of MMF estimators to suppress the interference of clutter \cite{Davis2007}, \cite{Stoica2007}, \cite{Ackroyd1973}.

The MMF estimator uses  a general real-valued sequence $\bm{x}$ instead of the phase sequence $\bm{s}$, and correlates the received sequence, giving
\begin{equation}\label{estimator2}
  \hat{h}_{0}=\frac{\bm{x}^{T}\bm{y}}{\bm{x}^{T}\bm{s}}=h_{0}+
  \sum^{N-1}_{n=1-N,n\neq0}h_{n}\frac{\bm{x}^{T}\textbf{J}_{n}\bm{s}}
  {\bm{x}^{T}\bm{s}}.
\end{equation}
The receiver optimizes $\bm{x}$ by maximizing the following formula:
\begin{equation}\label{MMF}
 \gamma_{\textrm{MMF}}=\frac{(\bm{x}^{T}\bm{s})^2}{\sum^{N-1}
 _{n=1-N,n\neq0}(\bm{x}^T\textbf{J}_{n}\bm{s})^2}.
\end{equation}

It has been shown in \cite{Stoica2007} that, given a phase code $\bm{s}$, the optimal sequence $\bm{x}$ that maximizes $\gamma_{\textrm{MMF}}$ is $\bm{x}^{*} = \textbf{R}^{-1}\bm{s}$,
where matrix $\textbf{R}$ is given by
\begin{equation}\label{matrix_R}
  \textbf{R}=\sum^{N-1}_{n=1-N,n\neq 0}(\bm{x}^{T}\bm{s}\bm{s}^{T}\textbf{J}_{n}^{T}).
\end{equation}
Substituting $\bm{x}^{*} = \textbf{R}^{-1}\bm{s}$ into (\ref{MMF}), we have
\begin{equation}\label{MMF2}
  \gamma_{\textrm{MMF}}=\bm{s}^{T}\textbf{R}^{-1}\bm{s}.
\end{equation}
Note that $\gamma_{\textrm{MMF}}$ only depends on the phase code $\bm{s}$. Hence, the objective for the design of the MMF estimator is then to discover a phase-code s that can maximize $\gamma_{\textrm{MMF}}$ in (\ref{MMF2}).

\subsection{HpGAN for Pulse Compression Radar}
In this section, we use HpGAN to search the phase sequences for pulse compression radar which maximize the metric $\mathcal{M}(\bm s)$. The metric $\mathcal{M}(\bm s)$ is defined as
\begin{equation}\label{metrc-radar}
  \mathcal{M}(\bm s) =\bm{s}^{T} \textbf{R}^{-1}\bm{s},
\end{equation}
where matrix $\textbf{R}$ is given by (\ref{matrix_R}).

Massive training data is a necessary condition for HpGAN. Unfortunately, there are no suitable mathematical constructions which can generate a large number of sequences with good metric $\mathcal{M}(\bm s)$ with an identical length. For this, we employed genetic algorithm (GA) to generate 250 different sequences as the training sequences of HpGAN. In the GA, we used random sequences as parents and then searched for sequences through crossover, mutation and selection operations where the fitness function is the metric $\mathcal{M}(\bm s)$. For the MMF estimator, these sequences which are generated by GA yielded SIR $\gamma_{\textrm{MMF}} \in[10,21]$.

For this application, the main architecture of HpGAN is similar to that of the network in Section III. Major differences in the specific design and parameter selection are summarized as follows:
\begin{itemize}
  \item In the encoder, we aim to increase the diversity of encoded sequences while ensuring high decoding accuracy. Therefore, the number of sample sequences in definition 1 is set to $P\leq 8$, because of the larger sequence length than that in Section III, whilst the bias vector is still set to $\textbf{b}\in [0,0.4]$.
  \item Since we need to convert the encoded two-dimensional sequences into one-dimensional and then feed it into the GAN, the length of the training sequences is $59\times59$. A simple perceptron model is difficult to effectively extract the characteristics of the training data. Therefore, we use a multi-layer perceptron as the generative model and discriminant model in HpGAN, in which each perceptron contains two hidden layers in this experiment. The parameters settings are listed in Table 2.
  \item In this experiment, our goal is not to generate sequences similar to the initial training sequences, but to generate sequences that are better than the initial sequences. However, training the network with only initial sequences cannot achieve our goal, because the principle of GAN is to generate data with similar characteristics to the training data. In this regard, the sequences that we generate during network training are better than the initial sequences, as new training sequences to update the initial sequences. Moreover, after every 2500 iterations (according to our experience, the model reaches equilibrium after 2500 iterations), we exploit the current model to generate a new sequence set $\mathcal{S}_{new}$ to update the initial training sequence set $\mathcal{S}_{init}$, i.e., $\mathcal{S}_{init}\leftarrow \mathcal{S}_{new}$.
\end{itemize}

\begin{table*}[ht!]
\caption{The parameters of HpGAN}
\centering
\begin{tabular}{c|c|c}
  \hline
  \textbf{Items} & \textbf{Parameters} & \textbf{Definition} \\
  \hline
  \multirow{4}*{Generative Model} & $I_{G}=100$ & Number of neurons in the input layer \\
  \cline{2-3}
    &$H^{1}_{G}=1024$ & Number of neurons in the first hidden layer \\
    \cline{2-3}
    &$H^{2}_{G}=1024$ & Number of neurons in the second hidden layer \\
  \cline{2-3}
    &$O_{G}=3481$ & Number of neurons in the output layer \\
  \hline
  \multirow{4}*{Discriminant Model} & $I_{D}=3481$ & Number of neurons in the input layer \\
  \cline{2-3}
    &$H^{1}_{D}=1024$ & Number of neurons in the first hidden layer \\
  \cline{2-3}
    &$H^{2}_{D}=1024$ & Number of neurons in the second hidden layer \\
  \cline{2-3}
    &$O_{D}=1$ & Number of neurons in the output layer \\
  \hline
  \multirow{2}*{Generative Model/Discriminant Model}
    &$\alpha=0.0001$ & Learning rate of the generative/discriminant model \\
  \cline{2-3}
    & $\hbox{batch}=100$ & Mini-batch size \\
  \hline
\end{tabular}
\end{table*}
\subsection{Performance Evaluation}
In the experiment, HpGAN was executed $2 \times 10^4$ iterations to train the model in order to generate better sequences. During training, the training set was updated a total of 7 times, specifically, the training set was updated every 2500 iterations. When the training set is updated for the $k$th time, we exploit the current generative model to generate a large amount of data, and select 100 different sequences as the new training set $\mathcal{S}_{new}$. Since the mean metric of the sequences in the initial training set $\mathcal{S}_{init}$ is 13 and we also consider the learning ability of HpGAN, the metric of sequence in the new training is set to $\mathcal{M}(\bm s)\geq 13+3\times k$, $\bm s \in \mathcal{S}_{init}$. As in Section III,  we calculate the loss values of the generative model and the discriminant model to observe the convergence of the network. To monitor the evolution of HpGAN, every 100 iterations, we evaluated the searching capability of HpGAN and record their mean metric $E[\mathcal{M}]$ and maximum metric $\hbox{max}[\mathcal{M}]$. We exploit the minimum metric $\hbox{max}[\mathcal{M}]$ and the mean metric $E[\mathcal{M}]$ in the generated sequence sets to reflect the search effect and search trend of HpGAN.
\begin{figure}[h]
  \subfigure[The loss values of the discriminator of HpGAN]{
    \begin{minipage}{8cm}
      \centering
      \includegraphics[width=3.0in]{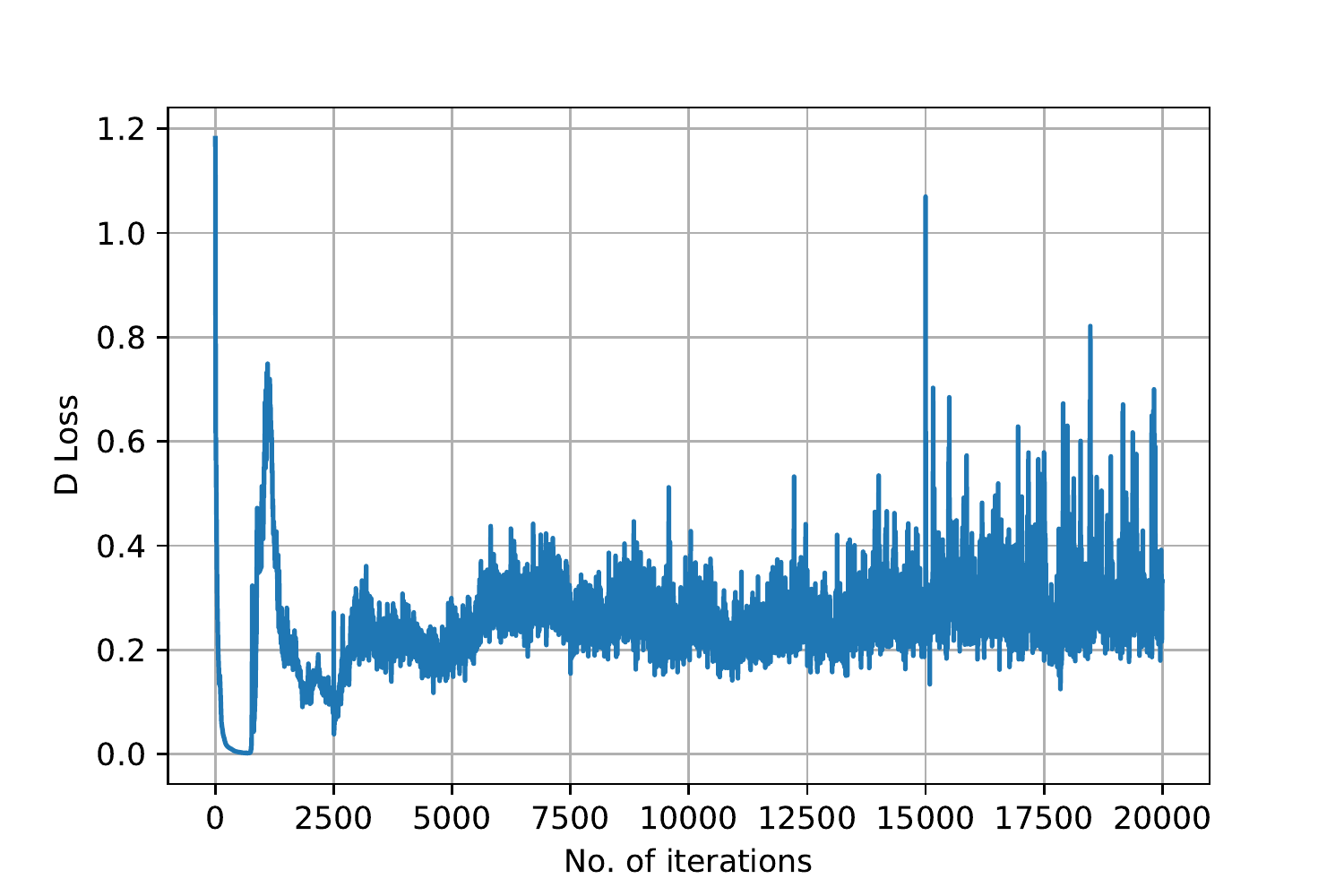}
    \end{minipage}}%

  \subfigure[The loss values of the generator of HpGAN]{
    \begin{minipage}{8cm}
      \centering
      \includegraphics[width=3.0in]{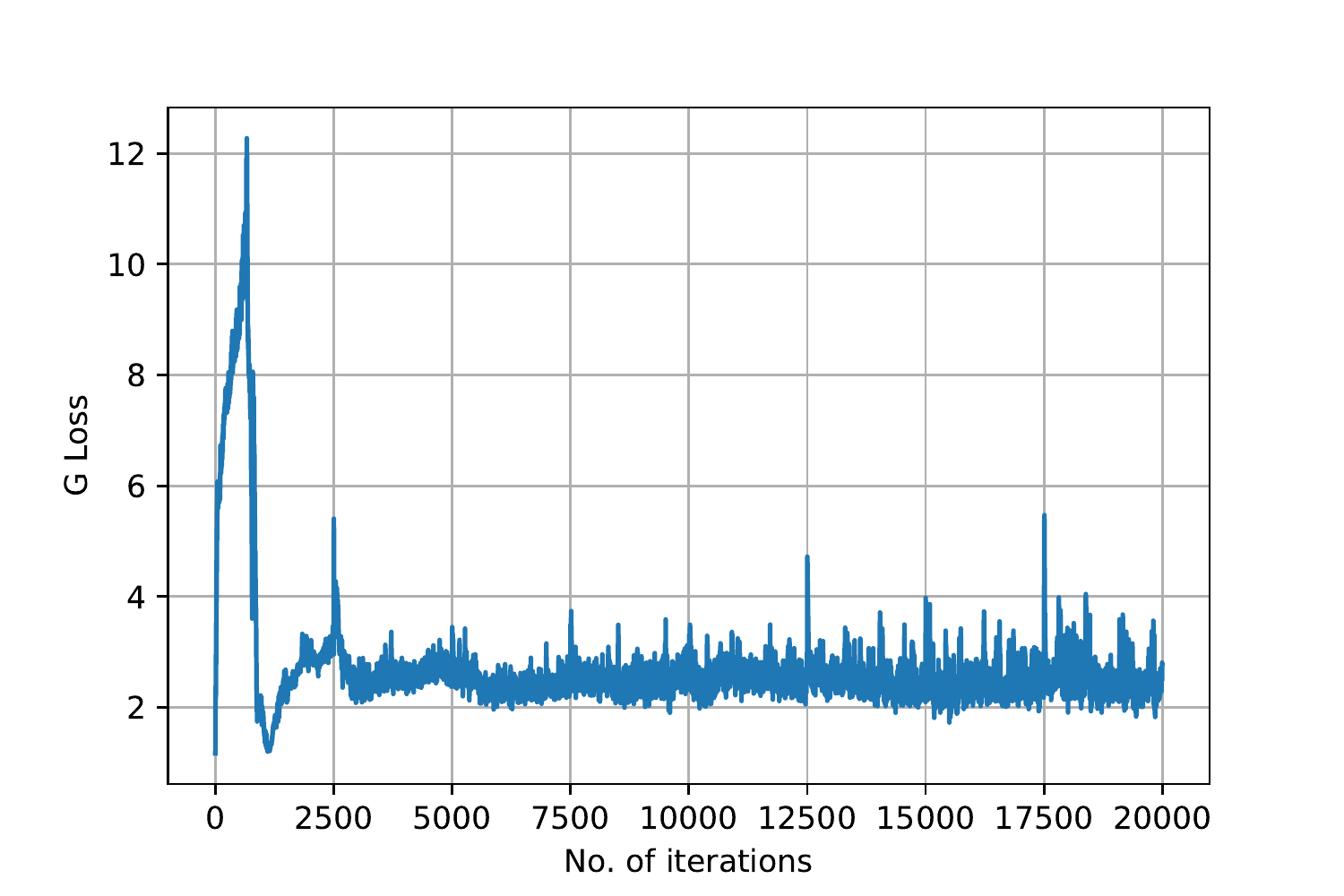}
    \end{minipage}}
  \caption{ The loss values of the generator and discriminator
of HpGAN. Each time the training set is updated, the loss functions of the model fluctuate and then gradually stabilizes.}
\end{figure}
\begin{figure}[h]
  \centering
  \includegraphics[width=3.0in]{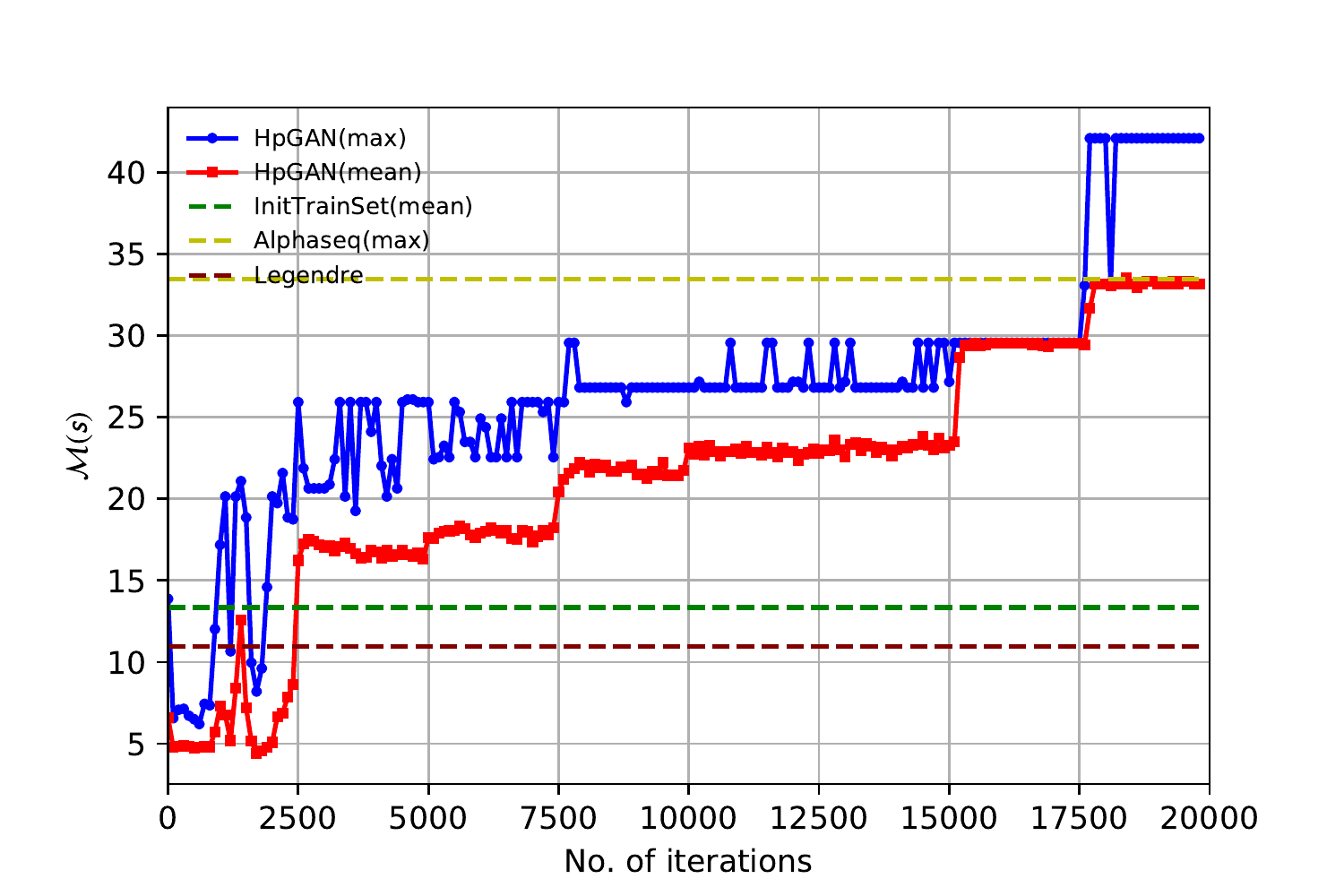}
  \caption{Training process of HpGAN to search a phase-coded sequence for pulse compression radar with the evolution curves of the mean metric $E(\mathcal{M})$ and the maximum metric $\hbox{max}(\mathcal{M})$.}
\end{figure}

From Fig. 8, we can see that after 2500 iterations, the loss functions of HpGAN gradually converges. Each time the training set is updated, the loss functions of the model fluctuate and then gradually stabilize showing how the model makes adjustment after the training set is updated.

The evolution curves of the mean metric $E(\mathcal{M})$ and maximum metric $\hbox{max}(\mathcal{M})$ with respect to the number of iterations are shown in Fig. 9. We can see that the overall trends of $E(\mathcal{M})$ and $\hbox{max}(\mathcal{M})$ is gradually increase with the number of iterations, especially after updating the training set, both $E(\mathcal{M})$ and $\hbox{max}(\mathcal{M})$ get greatly improved. The green dashed and yellow dashed lines represent the mean metric $E(\mathcal{M})_{\textrm{init}}$ of the initial training set and the maximum metric $\hbox{max}(\mathcal{M})_{\textrm{Alpha}}$ obtained by the AlphaSeq search, respectively. It can be seen from Fig. 9 that when the training set is updated for the first time, the sequences generated by HpGAN are better than the initial training set. After 10,000 iterations, HpGAN has found sequences whose metrics are better than $\hbox{max}(\mathcal{M})_{\textrm{Alpha}}$, and when the training set is updated for the last time, the mean metric of HpGAN generated sequences is close to $\hbox{max}(\mathcal{M})_{\textrm{Alpha}}$. This figure also demonstrates that HpGAN is an effective search tool, which can continuously improve the search ability as the training set is updated.

When HpGAN training is completed, we can exploit the generator to quickly generate sequences with linear complexity. After the 20000th iteration, we exploit the generator searches  a sequence with metric $\mathcal{M}(\bm{s}_{\textrm{HpGAN}})\approx 45.16$.
\begin{equation*}
\begin{aligned}
  \bm{s}_{\textrm{HpGAN}}&=[0 1 0 1 0 1 0 1 0 1 0 1 0 1 0 1 1 0 1 0 1 0 1 0 0 1 0 1 1 1 0 0 1 1 0 0 0 0 1 1\\
  &~~~~~~1 1 1 1 0 0 0 0 0 0 0 0 0 0 0 0 0 0 0].
\end{aligned}
\end{equation*}
Compared to the well-known Legendre sequence, $\bm{s}_{\textrm{HpGAN}}$ increases the signal-to-noise ratio of the MMF estimator in the pulse compression radar system by four times, and compared with the $\bm{s}_{\textrm{Alpha}}$ which is discovered by AlphaSeq, $\bm{s}_{\textrm{HpGAN}}$ improves the SIR by 11.71 at output of an MMF estimator.

\section{Conclusion}
In this paper, we proposed a novel algorithm for searching sequences based on GAN, which is called HpGAN. Unlike traditional heuristic algorithms and RL which may consume large amount of time, in the HpGAN training process, the network is updated inversely according to the loss function, avoiding the calculation of the metric function of the sequence.

We demonstrated the search ability of HpGAN through two applications. In the first application, we successfully found MOCCSs and optimal/sub-optimal OB-ZCPs, showing that HpGAN is a good supplement to the mathematical constructions. In the second application, based on the training set generated by the GA algorithm, HpGAN found a new sequence, which is far superior to the existing sequences for increasing the signal-to-noise ratio of the MMF estimator in the pulse compression radar system. Compared with the sequence discovered by AlphaSeq, the sequence found by HpGAN improves the SIR by 11.71 at output of an MMF estimator.

As a future work, more efficient GAN architecture (e.g., convolutional neural network) for sequences may be designed. Designing a more suitable encoder and decoder is an effective solution for searching longer sequences.

	\end{document}